%% file: survey_latex.tex
\documentclass[11pt]{article}

\usepackage[preprint]{acl}

\usepackage{fontspec}
\newfontfamily\arabicfont[Script=Arabic]{Amiri-Regular.ttf}[
    Path = Amiri/,
    BoldFont = Amiri-Bold.ttf,
    ItalicFont = Amiri-Italic.ttf,
    BoldItalicFont = Amiri-BoldItalic.ttf
]
\newfontfamily\symbolafont[
    Path = ./,
]{Symbola.ttf}

\usepackage{polyglossia}
\setdefaultlanguage{english}
\setotherlanguages{arabic,farsi,kurdish}
\newcommand{\AR}[1]{\textarabic{#1}}
\newcommand{\FR}[1]{\textfarsi{#1}}

\usepackage{latexsym}
\usepackage{amssymb}
\usepackage[T1]{fontenc}
\usepackage[utf8]{inputenc}
\usepackage{microtype}

\usepackage{inconsolata}
\usepackage{graphicx}
\usepackage[dvipsnames]{xcolor}
\usepackage[shortlabels]{enumitem}
\usepackage[disable]{todonotes}
\usepackage{booktabs}
\usepackage{multirow}
\usepackage{subcaption}
\usepackage{float}
\usepackage{pifont}
\usepackage{pdflscape}
\usepackage{longtable}
\usepackage[skip=0.5pt]{caption}
\usepackage{titlesec}
\titlespacing*{\section}{0pt}{4pt plus 1pt minus 1pt}{2pt}
\titlespacing*{\subsection}{0pt}{3pt}{1pt}

\usepackage{colortbl}

\newcommand{\graycell}[1]{%
\ifcase#1
#1%
\or\cellcolor{black!16}#1%
\or\cellcolor{black!22}#1%
\or\cellcolor{black!28}#1%
\or\cellcolor{black!34}#1%
\or\cellcolor{black!40}#1%
\or\cellcolor{black!46}\textcolor{white}{#1}%
\or\cellcolor{black!52}\textcolor{white}{#1}%
\or\cellcolor{black!58}\textcolor{white}{#1}%
\or\cellcolor{black!64}\textcolor{white}{#1}%
\or\cellcolor{black!70}\textcolor{white}{#1}%
\or\cellcolor{black!76}\textcolor{white}{#1}%
\or\cellcolor{black!82}\textcolor{white}{#1}%
\or\cellcolor{black!88}\textcolor{white}{#1}%
\else\cellcolor{black!92}\textcolor{white}{#1}%
\fi
}

\definecolor{darkViolet}{RGB}{123,50,148}
\definecolor{Violet}{RGB}{194,165,207}
\definecolor{beige}{RGB}{230,230,230} 
\definecolor{lightGreen}{RGB}{166,219,160} 
\definecolor{Green}{RGB}{0,136,55} 

\newcommand{\sameCountryCorrect}[1]{\cellcolor{darkViolet}\textcolor{white}{#1}}

\newcommand{\sameCountryIncorrect}[1]{\cellcolor{Violet}\textcolor{black}{#1}}

\newcommand{\diffCountryDiff}[1]{\cellcolor{Green}\textcolor{white}{#1}}

\newcommand{\diffCountrySomeSim}[1]{\cellcolor{lightGreen}\textcolor{black}{#1}}

\newcommand{\diffCountrySim}[1]{\cellcolor{beige}\textcolor{black}{#1}}

\makeatletter
\newcommand\newtag[2]{#1\def\@currentlabel{#1}\label{#2}}
\makeatother

\definecolor{LangEN}{HTML}{1B262C}
\definecolor{LangFR}{HTML}{0F4C75}
\definecolor{LangAR}{HTML}{3282B8}
\newcommand{\langEN}[1]{\textcolor{LangEN}{#1}}
\newcommand{\langFR}[1]{\textcolor{LangFR}{#1}}
\newcommand{\langAR}[1]{\textcolor{LangAR}{\AR{#1}}}
\newcommand{\captionBox}[2]{\raisebox{2pt}{\fcolorbox{black}{#1}{\rule{0pt}{1pt}\rule{1pt}{0pt}}}~#2}

\definecolor{a}{RGB}{127,60,8}
\definecolor{b}{RGB}{224,130,20}
\definecolor{c}{RGB}{254,224,182}
\definecolor{d}{RGB}{216,218,235}
\definecolor{e}{RGB}{128,115,172}

\definecolor{decreased_usage}{RGB}{239,158,60}
\definecolor{same_usage}{RGB}{232,233,241}
\definecolor{increased_usage}{RGB}{83,38,135}

\newcommand{\nparticipants}{189 }
\newcommand{\nfullparticipants}{170 }
\newcommand{\nFiveCountriesParticipants}{144 }
\newcommand{\lexiconWordsnosoapec}{210}
\newcommand{\lexiconWords}{\lexiconWordsnosoapec }
\newcommand{\lexiconWordsPerDialect}{$\geq40$ }
\newcommand{\Qsurveyperception}{\langEN{For you, Arabizi ...}}
\newcommand{\QEverUsedArabizi}{\langEN{Have you ever used Arabizi?}}
\newcommand{\QUsage}{\langEN{How is your use of Arabizi now compared to when you started texting on mobile phones?}}
\newcommand{\QResponseArabizi}{\langEN{If someone writes to you in Arabizi, would you more probably reply in Arabizi or in Arabic script?}}
\newcommand{\QResponseArabic}{\langEN{If someone writes to you in Arabic script, would you more probably reply in Arabizi or in Arabic script?}}
\newcommand{\QsurveyWhyPerception}{\langEN{Please explain why you think so.}}
\newcommand{\QsurveyComments}{\langEN{Are there any comments that you want to share with us before starting the second part?}}
\newcommand{\nVulgarSentences}{13 }
\newcommand{\nTransliteraionIssueSentences}{14 }
\newcommand{\nUniqueSentences}{189 }
\newcommand{\nTotalUniqueSentences}{203 }

\newcommand{\lexicon}{\textbf{\textit{VARabiziLex}}}
\newcommand{\parallelnadi}{\textbf{\textit{VARabiziCues}}}

\newcommand{\taha}{\todo[inline, color=cyan]}
\newcommand{\amr}[1]{\textcolor{black}{#1}}
\newcommand{\amro}[1]{\textcolor{black}{#1}}
\newcommand{\amroo}[1]{\textcolor{black}{#1}}
\usepackage[normalem]{ulem}
\usepackage{listings}
\usepackage{xcolor}  %

\title{\amr{Romanized Arabic} Across Dialects: Views, Usage Patterns, and Linguistic Variation}

\author{
 \textbf{Amr Keleg\textsuperscript{1,*}},
 \textbf{Ahmed Amine Ben Abdallah\textsuperscript{2}},
 \textbf{Taha Yassine\textsuperscript{3}},
 \textbf{Chadi Helwe\textsuperscript{4}},
\\
 \textbf{Imane Guellil\textsuperscript{5,*}},
 \textbf{Nedjma Ousidhoum\textsuperscript{6}}
\\
 \textsuperscript{1}MBZUAI,
 \textsuperscript{2}Cedar Rose,
 \textsuperscript{3}Independent Researcher,
 \textsuperscript{4}Lebanese American University,\\
 \textsuperscript{5}University of Birmingham,
 \textsuperscript{6}Cardiff University
\\
 \small{
   \textbf{Correspondence:} \href{mailto:amr.keleg@mbzuai.ac.ae}{amr.keleg@mbzuai.ac.ae}, \href{mailto: OusidhoumN@cardiff.ac.uk}{ OusidhoumN@cardiff.ac.uk}
 }
}

\begin{document}
\maketitle

\begin{abstract}
\amr{Arabizi refers to Arabic written in Latin script.}{\let\thefootnote\relax\footnotetext{* Work started while at the University of Edinburgh.}}
Although previous studies have shown that the prevalence and usage of Arabizi vary by factors such as region and age group, most NLP research on Arabic texts treats it as a temporary phenomenon resulting from limited technological support for the Arabic script.
In this work, we engage with Arabic speakers to collect insights on their perceptions and usage of Arabizi. 
We further examine writing norms among speakers of different dialects, focusing on Algerian, Egyptian, Lebanese, Moroccan, and Tunisian Arabic. To this end, we release two resources. First, we perform a character-level alignment of Arabic words to study inter- and intra-dialectal variation across these five dialects, based on words transliterated by the surveyed participants (\lexicon), finding systematic intra-dialectal regularity and inter-dialectal variation. Second, to study Arabic speakers' ability to identify this stylistic variation at the sentence-level, we build \parallelnadi, a manually curated parallel corpus of sentences written in Arabic script alongside multiple Arabizi transliterations, collected from speakers of the same five dialects. 
Our study presents the largest human-centered, cross-dialectal study of Arabizi's perceptions and practices to date.\footnote{\amroo{Our code and data are available at \href{https://github.com/AMR-KELEG/arabizi-across-dialects}{this link}.}}

\end{abstract}

\section{Introduction}

Arabizi, Franco Arabic, or Arabish\footnote{Hereafter, ``Arabizi'' denotes Romanized Arabic.} is a writing system that uses Latin letters and numerals to represent Arabic---including both Modern Standard Arabic~(MSA) and Dialectal Arabic~(DA) varieties \cite{darwish-2014-arabizi}. The system emerged out of necessity, as early phones and even smartphones did not properly support the Arabic script. Despite the widespread availability of Arabic keyboards today, Arabizi is still being used. It relies on the writers' intuitions in selecting Latin letters that best approximate the pronunciation of Arabic words, which is also variable, making it a non-standardized writing system.

\begin{figure}[t]
    \centering
    \vspace{3mm}
    \includegraphics[trim={0cm 1.25cm 0 0},clip]{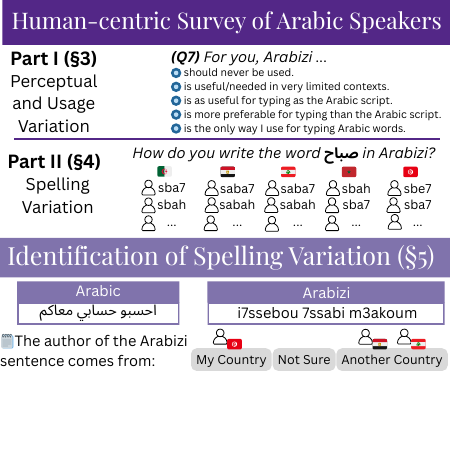}
    \caption{Overview of the paper's components to study the cross-dialectal variation in Arabizi.}
    \label{fig:placeholder}
    \vspace{-3mm}
\end{figure}

It is commonly believed that Arabizi is used primarily for DA \cite{al-badrashiny-etal-2014-automatic,adouane-etal-2016-romanized,tobaili-2016-arabizi}.
Unsurprisingly, most Arabizi data collected for NLP has focused on DA.
Moreover, 
prior research focused on country- or region-specific varieties (\citealp{bies-etal-2014-transliteration,cotterell2014algerian,tobaili-etal-2019-senzi,moudjari-etal-2020-algerian,fourati-etal-2021-introducing,riabi-etal-2023-enriching,wafa2024arabizi}; inter alia). \amro{Hence, findings would be mostly localized, making the study of cross-dialectal commonalities and variation underexplored.} Cross-dialectal studies could provide valuable insights into broader linguistic and social trends, supporting the development of more dialect-aware language technologies.

We address this gap by engaging with \nparticipants Arabic speakers from different countries. %
We examine their perspectives and usage of Arabizi, summarizing their diverse views. While we observe signs of decline in Arabizi use \amroo{among the surveyed participants from some countries} such as Egypt, a non-negligible proportion of participants report continued and frequent reliance on it. \amroo{Despite not having a statistically representative sample from the studied countries, our study still shows that speakers from different communities can have different views on Arabizi's utility. Hence, researchers need to acknowledge that their views on Arabizi's use are biased by the extent of their exposure to diverse speaking communities, and avoid generalizing their views and localized experiences to other speaking communities to which they are not exposed.}

To study the linguistic variation in Arabizi, we build two complementary resources.
For character- and word-level variation, we collect transliterations from speakers of the five national dialects most represented in previous studies (see \S\ref{sec:dialects_in_studies}): three from the ``Maghreb'' (the Western Arabic-speaking region---Algeria, Morocco, and Tunisia) and two from the ``Mashriq'' (the Eastern region---Egypt and Lebanon). \amr{We find systematic inter-dialectal variation in how some characters (e.g., \AR{جـ}, \AR{ذ}, \AR{ق}) are rendered in Arabizi, and non-negligible intra-dialectal regularity in how words are generally transliterated.} 
To examine Arabic speakers' ability to recognize this spelling variation, we further collect parallel transliterations of the same Arabic-script sentences from participants across the five countries. The sentences were carefully chosen to lack country-specific lexical terms, thereby requiring participants to rely on dialect-specific orthographic/phonological cues. We find that Arabic speakers are generally able to recognize transliterations by speakers of their dialect, in addition to a non-negligible ability to identify spelling cues linked to other dialects.

To our knowledge, this paper is the first cross-dialectal study of Arabizi on this scale. Our human-centric approach enables us to analyze the inter-dialectal variation in Arabizi on both perceptual and linguistic aspects. \amro{We discuss the views of a diverse set of Arabic speakers from different countries. Our study's results provide insights for researchers interested in building language technologies, particularly those intended to serve the linguistically diverse community of Arabic speakers and learners. This complements research which highlighted the safety issues in LLMs that are caused by disregarding Arabizi \cite{ghanim-etal-2024-jailbreaking}.}
Our contributions can be summarized as follows:
\begin{enumerate}[label=\arabic*., leftmargin=*, noitemsep, nolistsep]
\item The first cross-dialectal, human-centered study capturing diverse perceptions and usage patterns of Arabizi among \nparticipants Arabic speakers across five regional dialects.
\item The release of \lexicon, a \lexiconWordsnosoapec-word lexicon (\lexiconWordsPerDialect per dialect), with multiple Arabizi forms contributed by participants, \amro{the first of its kind. The lexicon enables} fine-grained analysis of spelling variation within and across dialects.
\item An expert analysis of native Arabic speakers' ability to identify dialectal cues in Arabizi text lacking distinctive lexical markers, using \parallelnadi , a parallel corpus of \nUniqueSentences Arabic-Arabizi sentences.
\end{enumerate}

\section{Background}
\label{sec:background}
In contrast to romanization schemes such as \citeposs{LDC2004L02}, which provide a one-to-one mapping between Arabic and Latin scripts, Arabizi relies on writers' intuitions about how to phonologically encode Arabic words using the Latin script \cite{al-badrashiny-etal-2014-automatic}.
In Arabizi, numerals are used in addition to Latin letters to represent Arabic phonemes that do not exist in English or French. Orthographic conventions sometimes influence the choice of numerals, such as mapping \AR{ع}~to~3 and \AR{ح}~to~7 \cite{yaghan2008}.

\subsection{Contrasting Linguistic Features of Arabizi}
\label{sec:arabizi_contrasting_features}

\paragraph{Sparsity and Regularity} The lack of standardization in Arabizi means that the same Arabic word can be represented by multiple Latin forms, resulting in \textit{sparsity}. For instance, \citet{cotterell2014algerian} found 69 Arabizi variants for \AR{إن شاء الله} in their corpus of Algerian Arabizi comments. However, the frequency of using \textit{ch} for \AR{ش} was an order of magnitude higher than that of \textit{sh}, a regularity also noted by \citet{abu2019writing}. \amroo{Moreover, \citet{Sullivan2017Writing} found that this regularity is more common for consonants than vowels.} We observe similar patterns of sparsity and regularity in our study (\S\ref{sec:spelling_variation}).

\paragraph{Verbosity and Ambiguity} Arabizi can more explicitly reflect dialectal variations in vowel usage than Arabic script, by representing both short and long vowels, resulting in \textit{verbosity} \cite[p.~6]{Salloum_2018}. In contrast, the Arabic script often omits short vowels, which are rendered as diacritics in formal writing \cite{habash2010introduction}. \amroo{However, this verbosity in Arabizi is not consistent}, since both short and long vowels can still be dropped in Arabizi. \amroo{For instance, the short vowels and the two long vowels \AR{ي} in \AR{حبيبي}~(my~love) can be written/dropped, forming a set of possible transliterations: 7abibi, 7bbi, 7abebe, and 7bibi \cite{tobaili-etal-2019-senzi}}.
Moreover, multiple Arabic letters---corresponding pairs of short and long vowels \amroo{(e.g., \AR{ُ} and \AR{و})}, or related phones \amroo{(e.g., the voiceless alveolar stop \AR{ت} and its emphatic counterpart \AR{ط})}---can be mapped to the same Latin letter. This makes some Arabizi words \textit{ambiguous}, especially when read out of context. For example, \textit{matar} could represent both \AR{مطار}~(airport) and \AR{مطر}~(rain) in multiple Arabic dialects \cite{Guellil2020}. \S\ref{sec:human_perception_of_spelling_variation} \amroo{empirically} investigates the impact of verbosity and ambiguity on the ability of Arabizi-fluent annotators to identify dialectal variations in Arabizi.

\subsection{Views of Arabizi}
The term \underline{Arab}\underline{\underline{izi}} originates from the fusion of two words: \textit{\underline{Arab}i (Arabic)} and \textit{Engl\underline{\underline{izi}} (English)} \cite{yaghan2008}. A more controversial interpretation, offered by \citet{Gonzalez-Quijano2014}, suggests that the term can be pronounced as ``Arab Easy,'' framing Arabizi as a more convenient writing system than the Arabic script. This naming controversy also reflects how Arabic speakers may hold diverse, strong, and sometimes polarized, views of Arabizi. Although a few studies have investigated users' perspectives and motives, these were generally limited in scope, focusing on individual countries \cite{yaghan2008,jaran2015use,wafa2024arabizi}, while other work primarily analyzed the pervasiveness of Arabizi usage \cite{tobaili-2016-arabizi,kashina2020case}. In contrast, our paper adopts a human-centric approach, allowing a diverse group of Arabic speakers from different regions and communities to share their views and report on their usage of Arabizi.

\section{Human-centered Study of Arabizi Views and Usage Patterns}

To comparatively study the behavioral and linguistic variation of Arabizi among Arabic speakers, we designed a survey of two parts: (1)\ a questionnaire to examine different views and usage patterns of Arabizi, and (2)\ a transliteration task to study variation in how words are rendered in Arabizi across dialects. See the survey questions in Appendix \S\ref{sec:survey_questions}.

\paragraph{\amro{Dissemination}} The survey was conducted online via Qualtrics from June 9 to August 31, 2025. We used snowball sampling by sharing the survey on social media (\textit{X}, \textit{LinkedIn}), personal networks, and mailing lists of research communities (\textit{SIGARAB}, \textit{Masakhane}, and \textit{North Africans in NLP}).  \amroo{Hence, the survey is expected to be more representative of the views of researchers, students, and AI enthusiasts.}
However, our sampling strategy minimizes spam responses, and mitigates the underrepresentation of Arabic speakers on reliable annotation platforms that can verify the participants' identities (e.g., Prolific).

\paragraph{\amro{Participation}} For inclusion, the survey questions were made available in Arabic, English, and French.
Since participation was voluntary, participants were not required to complete the entire survey. For the transliteration task, participants were asked to transliterate at least 10 words and could complete the survey over multiple sessions.
In total, \nparticipants{}responses were collected, of which \nfullparticipants{}participants completed both parts, with the median completion time being approximately 11 minutes. Participants represented diverse regions and spoke several varieties of DA, as elaborated in Appendix \ref{sec:survey_part1_detailed_questions}. The following subsections describe the first part of the survey, and present an analysis of its responses.

\begin{figure*}[tbph]
    \centering
     \begin{subfigure}[b]{0.98\textwidth} 
    \includegraphics[width=\textwidth, trim={0 0.26cm 0 0}, clip]{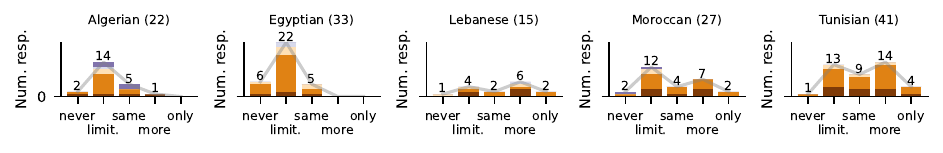}
     \caption{The distribution of the responses to \textit{(\ref{survey:perception}) \Qsurveyperception} \textbf{never}: \textit{should never be used}, \textbf{limit.}: \textit{is useful/needed in very limited contexts}, \textbf{same}: \textit{is as useful for typing as the Arabic script}, \textbf{more}: \textit{is more preferable for typing than the Arabic script}, \textbf{only}: \textit{is the only way I use for typing Arabic words}. Two Algerian participants provided free-form responses, which are not shown in the distribution: ``Sometimes easier for communication'', ``My keyboard is in Latin. I never change it''. \textbf{Note:}~Color encodes different age groups as follows:~\captionBox{a}{(<18)-24} \captionBox{b}{25-34} \captionBox{c}{35-44} \captionBox{d}{45-54} \captionBox{e}{55+}}
     \label{fig:perception_variation_country}
     \end{subfigure}

     \begin{subfigure}[b]{0.98\textwidth} 
    \includegraphics[width=\textwidth, trim={0 0.26cm 0 0}, clip]{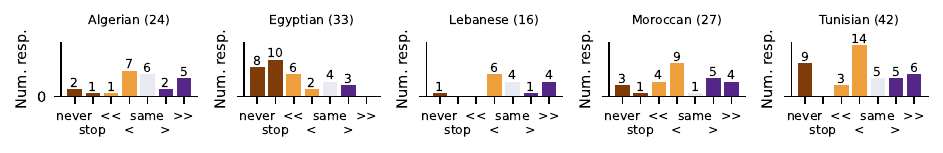}
    \caption{The distribution of the responses to \textit{(\ref{survey:usage}) \QUsage} \textbf{never}: answered \textit{no} to \textit{(\ref{survey:ever_used}) \QEverUsedArabizi}, \textbf{stop}: \textit{I stopped using Arabizi}, \textbf{<\hspace{0pt}<}: \textit{Much less now compared to when I started texting}, \textbf{<}: \textit{Somewhat less now compared to when I started texting}, \textbf{same}: \textit{About the same}, \textbf{>}: \textit{Somewhat more now compared to when I started texting}, \textbf{>\hspace{0pt}>}: \textit{Much more now compared to when I started texting}. \textbf{Note}:~Color encodes choices having the same direction;~\captionBox{a}{never used/stopped} ~\captionBox{decreased_usage}{decrease} ~\captionBox{same_usage}{same} ~\captionBox{increased_usage}{increase}.}
    \label{fig:usage_variation_country}
    \end{subfigure}

   \caption{The distributions of the perception (\ref{fig:perception_variation_country}) and usage of Arabizi (\ref{fig:usage_variation_country})  across the five studied countries. The two aspects are complementary. For instance, Algeria and Egypt have similar perception patterns, but completely different usage patterns. \textbf{Note:} Age Group is marginalized in \ref{fig:usage_variation_country}, given the skew shown in \ref{fig:perception_variation_country}. A few participants provided free-form answers to \ref{survey:perception}, hence, the number of responses reported for \ref{fig:perception_variation_country} is slightly less than that for \ref{fig:usage_variation_country}.}

\end{figure*}

\newpage

\subsection{Overview of the Views and Usage Patterns}
The survey's first part is a questionnaire designed to explore the participants' perceptions and usage of Arabizi. 
After collecting basic demographic information (age group, dialect, second language), the questionnaire addresses key factors, including:
\begin{itemize}[leftmargin=*, noitemsep, nolistsep]
\item Personal experience with Arabizi, such as, whether they have ever used it (\ref{survey:ever_used}), if and when they began texting on mobile phones, and how their Arabizi use has evolved over time (\ref{survey:usage}).
\item Contexts in which participants tend to use Arabizi, such as specific social media platforms.
\item Script preferences when responding to messages written in Arabizi~(\ref{survey:response_Arabizi}) and Arabic script~(\ref{survey:response_Arabic}).
\end{itemize}

We analyze variation in the perception of Arabizi (\ref{survey:perception}) and usage patterns (\ref{survey:usage}, \ref{survey:response_Arabizi}, and \ref{survey:response_Arabic}). Our analysis focuses on the five countries from which we obtained sufficient responses (\nFiveCountriesParticipants of the total \nparticipants responses). The distributions of responses to these questions, shown in Figures~\ref{fig:perception_variation_country},~\ref{fig:usage_variation_country},~\ref{fig:response_Arabizi_variation_country}, and~\ref{fig:response_Arabic_variation_country}, clearly demonstrate the variation present in both aspects, within and across countries.

\label{sec:cross_dialectal_percep}
\paragraph{Distribution of Perception Responses} We observe two distribution patterns (\autoref{fig:perception_variation_country}) in how Arabizi is perceived in the different countries.
The first pattern is \textbf{a bimodal distribution} observed in Lebanon, Morocco, and Tunisia. In these countries, some respondents consider Arabizi to have limited use (\textit{limit.}), while others express a preference for using it to type Arabic words (\textit{more}). A few respondents mentioned that they \textbf{\textit{only}} type Arabic words in Arabizi (\textit{n}=2 for both Lebanon and Morocco, and \textit{n}=4 for Tunisia).

The second pattern is \textbf{a unimodal distribution}, observed in Algeria and Egypt. In these cases, 13 out of 21 Algerian respondents and 22 out of 33 Egyptian respondents believe that Arabizi is useful only in limited contexts (\textit{limit.}). Meanwhile, six Egyptian and two Algerian participants think that Arabizi should never be used (\textit{never}). The latter observation suggests slightly greater acceptance of Arabizi in Algeria than in Egypt---an observation supported by the usage patterns discussed next. Check Appendix~\ref{sec:stat_sign} for Statistical significance.

\paragraph{Arabizi Usage Across Countries}
Based on the usage levels shown in \autoref{fig:usage_variation_country}, a distinct pattern emerges for Egypt, where ten respondents reported having stopped using Arabizi, compared to only one respondent (or none) in each of the other four countries. This could indicate stronger opposition to Arabizi in Egypt than in the other four countries, including Algeria, whose perception patterns otherwise resemble those of Egypt. This trend is also reflected in the user's preferred scripts in response to messages in Arabizi or in Arabic script as explained in \S\ref{sec:analysis_scripts} of the Appendix.

\paragraph{Decreased Usage of Arabizi?}
Although four to six respondents from all countries except Egypt indicated that their use of Arabizi has increased considerably since they began texting, there is a general tendency toward decreased usage of Arabizi. Across the five countries, more respondents reported using Arabizi less frequently (categories ``somewhat less'' \textit{<} and ``much less'' \textit{<\hspace{0pt}<}) than those reporting more frequent use (``somewhat more'' \textit{>} and ``much more'' \textit{>\hspace{0pt}>}), with a few participants reporting no clear change in their usage. Additionally, eight Egyptian and nine Tunisian respondents reported never using Arabizi, compared to three or fewer respondents in the other three countries. \amroo{We stress that we do not claim that these trends are representative of the entire populations of the studied countries. Our goal is to show that speakers from different communities could have different circumstances, needs, and views of Arabizi.}

\begin{table*}[h!]
    \centering
    \small
    \setlength{\tabcolsep}{2pt}
    \resizebox{\linewidth}{!}{%
    \begin{tabular}{p{3cm}p{8.5cm}c cccccc}
    \multicolumn{1}{c}{\multirow{2}{*}{\textbf{Theme}}} &
    \multicolumn{1}{c}{\multirow{2}{*}{\textbf{Subtheme}}} &
    \multirow{2}{*}{\textbf{N}} &
    \multicolumn{6}{c}{\textbf{(\ref{survey:perception}) \Qsurveyperception}} \\

    & & & never & limit & same & more & only & ... \\

    \midrule

    \multicolumn{9}{c}{\textbf{Anti-Arabizi Views}}\\
    \midrule

    \multirow{2}{3cm}{\newtag{(T1) Impact on Arabic as a language}{tag:T1}}
    & - Loss of Arabic proficiency and cultural identity
    & 15
    & \graycell{4} & \graycell{10} & 0 & 0 & \graycell{1} & 0\\

    & - Arabizi reduces the charm of Arabic
    & 4
    & 0 & \graycell{2} & 0 & \graycell{1} & 0 & \graycell{1}\\
    
    \midrule

    \multirow{4}{3cm}{\newtag{(T2) Weaknesses of Arabizi as a writing system}{tag:T2}}
    & - Only understood by its users (e.g., excluding older generations)
    & 4
    & \graycell{1} & \graycell{2} & 0 & 0 & \graycell{1} & 0\\

    & - Hard to read, when written by speakers of other dialects
    & 4
    & 0 & \graycell{4} & 0 & 0 & 0 & 0\\

    & - Arabizi is not Standardized
    & 3
    & \graycell{2} & \graycell{1} & 0 & 0 & 0 & 0\\

    & - Arabizi can not be used to type Arabic technical terms
    & 2
    & 0 & 0 & \graycell{1} & 0 & 0 & \graycell{1}\\

    \midrule

    \multirow{3}{3cm}{\newtag{(T3) Perceived absence of need for Arabizi}{tag:T3}}
    & - Arabic Keyboards are everywhere
    & 4
    & \graycell{4} & 0 & 0 & 0 & 0 & 0\\

    & - Arabic should be written in Arabic script
    & 3
    & \graycell{2} & \graycell{1} & 0 & 0 & 0 & 0\\

    & - Arabizi is Annoying
    & 2
    & \graycell{2} & 0 & 0 & 0 & 0 & 0\\

    \midrule
    \multicolumn{9}{c}{\textbf{Pro-Arabizi Views}}\\
    \midrule

    \multirow{4}{3cm}{\newtag{(T4) Inadequacy of support for Arabic}{tag:T4}}
    & - Only useful when Arabic keyboards are not available
    & 8
    & 0 & \graycell{6} & \graycell{1} & 0 & 0 & \graycell{1}\\

    & - Handy on computers, for lack of autocomplete
    & 4
    & 0 & \graycell{1} & \graycell{1} & \graycell{2} & 0 & 0\\

    & - Software issues (e.g., misalignment) for Arabic/mixed text
    & 3
    & \graycell{1} & \graycell{2} & 0 & 0 & 0 & 0\\

    & - Keyboards are generally designed for the Latin script
    & 2
    & 0 & 0 & 0 & \graycell{1} & 0 & \graycell{1}\\

    \midrule

    \multirow{7}{3cm}{\newtag{(T5) Convenience of using Arabizi}{tag:T5}}
    & - Typing in Arabizi is faster (a feature needed for chatting)
    & 13
    & 0 & \graycell{6} & \graycell{2} & \graycell{4} & \graycell{1} & 0\\

    & - Easier when code-switching (no need to change keyboards)
    & 11
    & 0 & \graycell{6} & \graycell{2} & \graycell{2} & 0 & \graycell{1}\\

    & - Arabizi is useful in informal texting
    & 10
    & 0 & \graycell{7} & \graycell{1} & 0 & \graycell{1} & \graycell{1}\\

    & - Arabizi is easier to use
    & 4
    & 0 & \graycell{1} & 0 & \graycell{1} & \graycell{1} & \graycell{1}\\

    & - I am used to it, and I feel guilty about that!
    & 3
    & 0 & \graycell{2} & 0 & \graycell{1} & 0 & 0\\

    & - Arabizi is only useful for convenience
    & 2
    & 0 & \graycell{2} & 0 & 0 & 0 & 0\\

    & - Arabizi blends better with slang, emojis, abbreviations
    & 1
    & 0 & 0 & 0 & \graycell{1} & 0 & 0\\

    \midrule

    \multirow{2}{3cm}{\newtag{(T6) Ignorance of Arabic script}{tag:T6}}
    & - Some Arabs (e.g., diaspora)/non-natives can not read Arabic script
    & 8
    & 0 & \graycell{5} & \graycell{3} & 0 & 0 & 0\\

    & - Arabizi is useful for Arabic learners
    & 5
    & 0 & \graycell{5} & 0 & 0 & 0 & 0\\

    \midrule

    \multirow{4}{3cm}{\newtag{(T7) Weaknesses of Arabic script as a writing system}{tag:T7}}
    & - Dropping diacritics makes some Arabic script text confusing
    & 2
    & 0 & 0 & 0 & \graycell{1} & 0 & \graycell{1}\\

    & - Arabizi is more faithful to phonetics
    & 1
    & 0 & 0 & 0 & \graycell{1} & 0 & 0\\

    & - No spelling rules in Arabizi (i.e., less stress on correctness)
    & 1
    & 0 & 0 & 0 & \graycell{1} & 0 & 0\\

    & - Arabic script is difficult to read on screens
    & 1
    & 0 & \graycell{1} & 0 & 0 & 0 & 0\\

    \midrule

    \multirow{3}{3cm}{\newtag{(T8) Linguistic needs of North African dialects}{tag:T8}}
    & - My dialect has loanwords, blending better in Arabizi
    & 2
    & 0 & \graycell{1} & \graycell{1} & 0 & 0 & 0\\

    & - Some Arabic keyboards do not include letters in my dialect like \textkurdish{ڤ}
    & 1
    & 0 & 0 & 0 & \graycell{1} & 0 & 0\\

    & - Arabizi is better suited for dialects that have no standard forms
    & 1
    & 0 & 0 & 0 & \graycell{1} & 0 & 0\\

    \bottomrule
    \end{tabular}%
    }

    \caption{The main themes and subthemes identified by qualitatively analyzing the comments provided by 91 participants. Note that a participant's comment may be linked to multiple subthemes. \textbf{N} is the total number of users mentioning a specific subtheme, and the subsequent columns indicate the distribution of these users according to \textit{(\ref{survey:perception}) \Qsurveyperception} \textbf{never}: \textit{should never be used}, \textbf{limit.}: \textit{is useful/needed in very limited contexts}, \textbf{same}: \textit{is as useful for typing as the Arabic script}, \textbf{more}: \textit{is more preferable for typing than the Arabic script}, \textbf{only}: \textit{is the only way I use for typing Arabic words}, \textbf{...}: \textit{free form responses}.}
    \label{tab:qualitative_analysis_themes}
\end{table*}

\subsection{Understanding Behavioral Patterns}
\amro{We used an affinity diagram \cite{HOLTZBLATT2017127}
to qualitatively analyze the free-form responses provided by 91 participants to (\ref{survey:why_perception}) \amroo{\textit{\QsurveyWhyPerception}} and (\ref{survey:comments}) \amroo{\textit{\QsurveyComments}}. One author identified subthemes, grouped them into main themes, and another author verified the themes and adjusted them if needed. \autoref{tab:qualitative_analysis_themes} reports the main themes, their subthemes, and respective frequencies (N).} \amroo{Comments covering multiple subthemes are attributed to each independently (i.e., counted multiple times).}

We present the key themes categorized by the participants' answers to (\ref{survey:perception}) \textit{\Qsurveyperception}

\paragraph{Arabizi Should Never Be Used}
The most prominent theme (8/16) is \ref{tag:T3}, focusing on the abundance of Arabic keyboards, and the obligation of writing Arabic in Arabic script. This is followed by \ref{tag:T1} (4/16), where fears of loss of identity and language proficiency are mentioned, and \ref{tag:T2} (3/16), highlighting the lack of standardization and unintelligibility by people who do not use it.

\paragraph{Arabizi is Useful in Limited Contexts} \hyperref[tag:T1]{(T1)} and \hyperref[tag:T2]{(T2)} were also mentioned in (12/60) and (7/60) comments respectively. However, Arabizi is also viewed as a pragmatic, context-bound workaround for: \ref{tag:T5} (24/60) like typing speed, \ref{tag:T6} (10/60) especially for native and non-native speakers who can not read the script, and \ref{tag:T4} (9/60).

\begin{table*}[!bhpt]
    \small
    \centering
    \setlength{\tabcolsep}{0.5pt} %
    \begin{tabular}{l @{\hskip 5pt} c @{\hskip 5pt} p{12cm} c}
        \textbf{Dialect} & \textbf{Word} & \textbf{Transliterations (Freqencies)} & \textbf{Mode/Total}\\
        \midrule
        \multirow{2}{*}{\textbf{Algeria}} & \textbf{\AR{رقيق}} & \textbf{r9i9~(11)} rgig~(2) r'qiq~(1) re9i9~(1) rekik~(1) reqiq~(1) rghigh~(1) rqiq~(1) rquiq~(1) rqyq~(1) & \textbf{11/20}\\ 
        & \textbf{\AR{قصبر}} & {9osbor~(5)} 9osber~(2) kesbar~(2) kosbor~(2) qosbor~(2) 9esber~(1) 9esbor~(1) 9osbar~(1) 9ssber~(1) gousbr~(1) kassbar~(1) qosbar~(1) & 5/21 \\
        \midrule
        \multirow{2}{*}{\textbf{Egypt}} & \textbf{\AR{قانون}} & \textbf{qanoon~(13)} qanon~(5) kanoon~(3) kanon~(2) qanun~(1) qanuun~(1) & \textbf{13/25}\\ 
        & \textbf{\AR{بتقول}} & {bet2ol~(8)} bet2ool~(7) bt2ol~(4) bt2ool~(2) bet'ool~(1) bet2ul~(1) betqool~(1) bt2oul~(1) & 8/25\\
        \midrule
        \multirow{2}{*}{\textbf{Lebanon}} & \textbf{\AR{رقيق}} & \textbf{r2i2~(8)} rakik~(3) r2e2~(1) ra2i2~(1) rkik~(1) & \textbf{8/14}\\ 
        & \textbf{\AR{قديم}} & \textbf{adim~(9)} 2adim~(4) kadeem~(1) & \textbf{9/14}\\ 
        \midrule
        \multirow{2}{*}{\textbf{Morocco}} & \textbf{\AR{رقيق}} & \textbf{r9i9~(14)} rqiq~(5) raquique~(1) re9i9~(1) & \textbf{14/21}\\ 
        & \textbf{\AR{قال}} & 9al~(9) gal~(8) qal~(1) qala~(1) qual~(1) & 9/20\\ 
        \midrule
        \multirow{1}{*}{\textbf{Tunisia}} & \textbf{\AR{قديم}} & \textbf{9dim~(18)} gdim~(7) kdim~(2) 9adim~(1) gdime~(1) & \textbf{18/29}\\ 
        \bottomrule
    \end{tabular}
    \caption{\textcolor{black}{Some words from \lexicon\  representative of} \AR{ق}, \textcolor{black}{with the participants' transliterations and respective frequencies. Despite being diverse, the most common (mode) transliteration shows signs of regularity for most words. \textbf{Note}: the most common transliteration is bolded if its frequency of usage exceeds 50\%.}}
    \label{table:lexicon_qaf}
\vspace*{-0.3mm}
\end{table*}

\paragraph{Arabizi is as Useful as Arabic Script}
Unlike the previous two groups, proponents of this view do not share concerns about Arabizi's threat to Arabic \hyperref[tag:T1]{(T1)} or mention the weaknesses of Arabizi \hyperref[tag:T2]{(T2)}. They only focus on the utility of Arabizi: \hyperref[tag:T5]{(T5)} and \hyperref[tag:T6]{(T6)} by (5/12) and (3/12), with (2/12) also referring to the lack of support for Arabic \hyperref[tag:T4]{(T4)}.

\paragraph{Arabizi is More Preferable than Arabic Script} Again, \ref{tag:T5} is the most prominent theme (9/18), as indicated by the following quote:
\textit{``I would have developed speed typing in Arabic had I owned an Arabic keyboard, but since I live/work abroad, I'm using tools like Yamli to help me type in Arabic, and now I use ChatGPT, which sometimes gets it right.''} Distinctively, a few participants referred to \ref{tag:T7} (3/18) and \ref{tag:T8} (2/18).

\paragraph{Arabizi is the Only Way to Type Arabic}
\ref{tag:T5} is mentioned by (4/5) participants. Notably, two comments referred to the drawbacks of using Arabizi 
\ref{tag:T1} and \ref{tag:T2}
, despite the fact that they only use Arabizi to type Arabic.

\section{Human-Centered Resource Creation for Studying Arabizi Spelling Variation}
\label{sec:spelling_variation}
The second part of the survey is designed to collect transliterations of Arabic words into Arabizi from speakers of the five countries. These transliterations enable us to investigate spelling regularities and variations at both the character and word levels.

\paragraph{Choice of Lexical Entities to Be Transliterated} \todo[color=green]{Revalidate the words that could not be aligned.}
When designing the survey, native speakers of Algerian, Egyptian, Lebanese, Moroccan, and Tunisian Arabic proposed sets of colloquial words commonly used in their respective countries. The sets were later refined and selected for inclusion in the questionnaire. For example, multiple representative words were included for the same letter when different mappings were expected based on phonological context or character position (e.g., the rendering of \AR{ﺓ} in \AR{شمعة} vs.\ \AR{حديقة الورد} in Lebanese Arabic, and \AR{ق} in \AR{قانون} vs.\ \AR{بتقول} in Egyptian Arabic as shown in \autoref{table:lexicon_qaf}). Characters such as \AR{ت} \textit{(t)} and \AR{ر} \textit{(r)}, which typically have one-to-one mappings in Arabizi, were not explicitly assigned to specific words but were expected to appear naturally within other words.
The words were shuffled differently for each participant to allow partial completion.

The resulting lexicon (\lexicon), listed in \S\ref{sec:lexicon_freq}, contains 45, 42, 41, 42, and 40 words for Algerian, Egyptian, Lebanese, Moroccan, and Tunisian Arabic, respectively, with corresponding average numbers of transliterations per word of 18.2, 24.6, 13.3, 19.5, and 28.8. Interestingly, the number of unique transliterations per word is 6.6, 5.3, 3.8, 5.1, and 6.2, respectively. Hence, signs of regularity exist (\S\ref{sec:arabizi_contrasting_features}), where a specific transliteration is commonly preferred, as shown in \autoref{table:lexicon_qaf}.

\begin{table*}[tbph]
\tiny
\centering
\setlength{\tabcolsep}{0.5pt} %
\begin{tabular}{l @{\hskip 10pt} p{2cm} p{2.5cm} p{2.5cm} p{2.1cm} p{1.8cm} p{3cm}}
 \textbf{Dialect} & \multicolumn{1}{c}{\textbf{\AR{خ}}} & \multicolumn{1}{c}{\textbf{\AR{غ}}} & \multicolumn{1}{c}{\textbf{\AR{ه}}} & \multicolumn{1}{c}{\textbf{\AR{ر}}} & \multicolumn{1}{c}{\textbf{\AR{ص}}} & \multicolumn{1}{c}{\textbf{\textkurdish{ڤ}}} \\
\midrule

\textbf{Algeria} & 
\textcolor{black}{kh~(20)} \textcolor{black}{5~(1)} & 
\textcolor{black}{gh~(19)} & 
\textcolor{black}{h~(34)} \textbf{8~(4)} \textcolor{black}{ah~(1)} & 
\textcolor{black}{r~(203)} \textbf{rr~(14)} & 
\textcolor{black}{s~(62)} \textbf{ss~(17)} & 
\textbf{g~(30)} \textbf{gh~(3)} \textbf{9~(2)} \textbf{f~(1)} \\

\textbf{Egypt} & 
\textcolor{black}{5~(17)} \textcolor{black}{kh~(8)} & 
\textcolor{black}{gh~(14)} \textcolor{black}{3'~(5)} \textcolor{black}{8~(5)} & 
\textcolor{black}{h~(46)} & 
\textcolor{black}{r~(215)} & 
\textcolor{black}{s~(49)} & 
\textcolor{black}{v~(24)} \\

\textbf{Lebanon} & 
\textcolor{black}{kh~(16)} \textcolor{black}{5~(12)} & 
\textcolor{black}{gh~(10)} \textcolor{black}{8~(5)} & 
\textcolor{black}{h~(28)} & 
\textcolor{black}{r~(152)} & 
\textcolor{black}{s~(29)} & 
\textcolor{black}{v~(14)} \\

\textbf{Morocco} & 
\textcolor{black}{kh~(77)} & 
\textcolor{black}{gh~(19)} & 
\textcolor{black}{h~(41)} \textbf{8~(2)} & 
\textcolor{black}{r~(300)} \textbf{rr~(15)} & 
\textcolor{black}{s~(29)} \textbf{ss~(16)} & 
\textcolor{black}{v~(21)} \textbf{9~(1)} \\

\textbf{Tunisia} & 
\textcolor{black}{5~(29)} \textcolor{black}{kh~(28)} & 
\textcolor{black}{gh~(21)} \textcolor{black}{8~(9)} & 
\textcolor{black}{h~(113)} \textcolor{black}{eh~(1)} & 
\textcolor{black}{r~(219)} & 
\textcolor{black}{s~(55)} & 
\textbf{g~(28)} \textbf{gh~(1)} \textbf{9~(1)} \\
\bottomrule
\end{tabular}
\caption{\amr{Latin equivalents of 6 characters, aggregated across different words and participants (See full table~\autoref{table:letter_mappings}). Mappings' frequencies are shown (between parentheses). } \textbf{Note:} Mappings not in the \protect{\href{https://en.wikipedia.org/wiki/Arabic_chat_alphabet\#Comparison_table}{Wikipedia table}} are \textbf{in bold}.}
\label{table:sub_letter_mappings_transposed}
\vspace*{-0.3mm}
\end{table*}

\todo[color=green]{
    >> No. of unique transliterations vs no. of transliterations for each word - Regularity against Sparsity in Arabizi (Recheck after \autoref{tab:part2_words_by_dialect} is updated)
}
    
\todo[color=green]{
    >> No. of transliterations that are shorter than the Arabic words (i.e., the transliteration is more ambiguous than the Arabic word) - Recheck after \autoref{tab:part2_words_by_dialect} is updated.
}

\vspace*{-0.3mm}

\paragraph{Transliteration Alignment}
To align the Arabic words of \lexicon\  with the collected transliterations, we developed a matching algorithm using a seed mapping table between Arabic and Latin letters, extracted from \href{https://en.wikipedia.org/wiki/Arabic_chat_alphabet\#Comparison_table}{Wikipedia's Arabic Chat page}. We then manually validated the mappings to ensure their correctness. Out of 1,390 unique transliterations, only 222 were affected by the manual validation, mostly because of how consecutive vowels are mapped.

\vspace*{-0.3em}

\paragraph{Analyzing the Character-level Mappings} In addition to identifying new letter mappings between Arabic and Latin scripts, the contrastive structure of \autoref{table:sub_letter_mappings_transposed} reveals notable inter-dialectal variations (see \S\ref{sec:spelling_variation_extended} and \autoref{table:letter_mappings} for a further discussion):

\begin{itemize}[leftmargin=*, noitemsep, nolistsep, topsep=0pt]
\item \textit{5} is used for \AR{خ} in Egyptian, Lebanese, and Tunisian transliterations, but only once in the Algerian data and never in Moroccan.
\item \textit{3'} is used exclusively in Egyptian transliterations to represent \AR{غ}. The numeral \textit{8} corresponds to \AR{غ} in Egypt, Lebanon, and Tunisia, but represents \AR{ه} in Algerian and Moroccan Arabizi.
\item Consonants such as \AR{ر} and \AR{ص} are often doubled (\textit{rr} and \textit{ss}, respectively) in Algeria and Morocco, a sign of the tendency to explicitly mark gemination ``\AR{تشديد}''~\cite{saadane-habash-2015-conventional}.
\item \textkurdish{ڤ} is rendered variably in Arabizi, reflecting inter-dialectal phonemic variation---as \textit{v} in Egyptian, Lebanese, and Moroccan Arabizi; and as \textit{g}, \textit{gh} in Algerian and Tunisian. Distinctively, Tunisians uniquely represent~\textit{v}~using~\textkurdish{ڥ}.
\item Many vowels (\textit{a, e, i, o, u}) align with an empty string ($\epsilon$) in the Arabic script, reflecting the common omission of short vowels in Arabic orthography, which are typically encoded in Arabizi.
\end{itemize}

\todo[color=cyan]{Discuss the variability in spelling by individual participants. E.g. \AR{عملولو} transliterated to 3aml\textbf{ou}l\textbf{o} (sample 1105 Moroccan participants).}

\section{A Parallel Corpus of Sentence-Level Arabizi Variations}
\label{sec:human_perception_of_spelling_variation}
\amr{This section investigates the salience of the character and word spelling variation found in \S\ref{sec:spelling_variation}. More specifically, we analyze whether Arabic speakers\amroo{---who are Arabizi-fluent---}can rely on this spelling variation to distinguish between the styles of speakers of different dialects. To this end, we constructed \parallelnadi, a dataset of parallel sentences transliterated into Arabizi by speakers of Algerian, Egyptian, Lebanese, Moroccan, and Tunisian Arabic. Annotators from the five countries were presented with sets of parallel Arabizi transliterations of the same sentence, and tasked to identify if each transliteration could have been written by someone from their own country.}

\begin{table*}[tbhp]
    \centering
    \small
    \input{exp2_results.tex}
    \caption{The acceptability judgments by 10 participants (columns). Each row represents a different set of transliterations, provided by a single participant. \textit{\textbf{N}} represents the total number of transliterations of each split (the same per country). \textbf{\textit{N\textsubscript{Com}}} represents the number of transliterations of each split, whose Arabic-script counterparts are considered valid in the column country's dialect. The values \checkmark/?/\ding{55} represent the number of samples that were judged as \textit{My Country}/\textit{Not Sure}/\textit{Another country}, respectively.\\ \textbf{Note 1:} \captionBox{darkViolet} cells on the diagonal mean that 90\% of the samples were correctly labeled as \textit{My Country}, \captionBox{Violet} otherwise.\\
    \textbf{Note 2:} For off-diagonal cells, \captionBox{Green} (the majority) indicates some ability to identify cues of other dialects, whereby >50\% of the samples were judged as \textit{Another country}, \captionBox{beige} indicates a degree of similarity between two countries, whereby >50\% of the samples were judged as \textit{My Country}, and \captionBox{lightGreen} indicates a degree of indecisiveness.}
\label{table:annotation_results}
\vspace*{-0.3pt}
\end{table*}

\subsection{Data Collection for the Parallel Corpus}
We built upon the MLADI dataset \cite{abdul-mageed-etal-2024-nadi,keleg-etal-2025-revisiting}, which contains 1,120 tweets (mostly in dialectal Arabic) evenly representing 14 Arabic-speaking countries. The dataset includes manually sourced annotations for tweet validity and the Arabic Level of Dialectness (ALDi) \cite{keleg-etal-2023-aldi} for 11 countries, including Algeria, Egypt, Morocco, and Tunisia. We collected labels for Lebanon as described in \S\ref{sec:lebanese_annotations}.
\vspace*{-0.3em}

\paragraph{Samples Selection}
We selected sentences that (1)\ are valid in at least two of the five considered countries and (2)\ have respective country-level average ALDi scores $\geq 0.33$, which correspond to \textit{Formal Colloquial} (ALDi~=~0.33), \textit{Ordinary Colloquial} (ALDi~=~0.67), or \textit{Informal Colloquial} (ALDi~=~1). Sentences with ALDi~<~0.33 are expected to be in MSA, which were discarded because their Arabizi transliterations would likely not be natural, except in rare cases such as prayers in MSA used in colloquial contexts. We filtered out \nVulgarSentences sentences containing expletives or inappropriate language, resulting in a total of \nTotalUniqueSentences sentences.

\paragraph{Samples Transliteration} 
Two participants from each country provided independent transliterations of the Arabic-script sentences valid in their respective dialects.\
\todo{Add the guidelines to the Appendix} 
We applied the matching algorithm described in \S\ref{sec:spelling_variation} to identify potential transliteration errors. We manually reviewed the transliterations, and discarded \nTransliteraionIssueSentences sentences that have minor typos in some transliterations. Hence, our parallel corpus (\parallelnadi) has a total of \nUniqueSentences sentences.

\vspace*{-0.3em}

\subsection{Annotating Parallel Arabizi Sentences}
\label{sec:experiment}

Annotators were presented with an Arabic-script sentence alongside a set of four or fewer possible transliterations, and were asked to assign each transliteration to one of three categories: \textit{My~Country}, \textit{Not~Sure}, and \textit{Another~Country} (\checkmark/?/\ding{55} in \autoref{table:annotation_results}), answering the question: \textbf{\textit{The author of the Arabizi (Franco~Arabic) sentence comes from?}} (see \autoref{fig:annotation_interface}). 
To minimize the cognitive load, the transliterations of sentences with more than 4 were split into multiple non-overlapping groups and presented independently. Consequently, each group presented could contain two, one, or none of the transliterations from a particular country \textit{C}, reducing any potential annotation biases or shortcuts. The same 10 participants who provided the transliterations acted as annotators for this experiment.

\paragraph{Arabizi Sentence Annotation Results} 
We group the transliterations provided by a single participant and analyze the judgments for each participant's transliterations (two per country) in \autoref{table:annotation_results}.
For each participant's transliterations, annotators only judged the subset of transliterations whose Arabic-script counterparts were valid in their respective country-level dialects. We observe the following:

\begin{itemize}[leftmargin=*, noitemsep, nolistsep]
    \item >90\% of same-country transliterations were \textcolor{black}{correctly categorized}, except for the Tunisian and Algerian annotators, who showed some hesitation to accept transliterations of the other participant from their country.
    
    \item On judging other-country transliterations, the participants were in general able to identify spelling cues and patterns indicative of those countries for which >50\% of the samples, labeling them as \textit{Another Country} (see the \captionBox{Green} cells).

    \item Variation in the judgments provided by participants from the same country exists, whereby one annotator may be more confident in selecting \textit{Another Country}, while the other chooses \textit{Not Sure} or even \textit{My Country}. This pattern is systematically observed among Moroccan annotators and, to a lesser extent, among Tunisian annotators.
\end{itemize}

In summary, we provide the first quantitative study which shows that Arabic speakers, fluent in Arabizi, are often more capable of guessing the dialect of a sentence in Arabizi than in Arabic script. \amroo{Future work could investigate the specific cues that the participants rely on when making their decision (e.g., rendering of short vowels, cross-dialectal variation in the choices of Latin letters and numerals).}

\todo[color=cyan]{Maybe we should check how sentences containing geographical clues were classified. For example, were the ones containing something clearly related to Algeria all accepted by Algerians and rejected by the others?}

\todo[color=green]{
\AR{الذبان فايق علي} in Tunisian. Do Tunisians have issues with adding any of the transliterations to "My Country"?
}

\section{Discussion}
\paragraph{Change is the Only Constant} 
We observe a general trend of declining Arabizi usage, with some users reporting that they have completely stopped using it, as shown in \autoref{fig:usage_variation_country}. This decline is most pronounced in Egypt, followed by Tunisia and Morocco. This observation is consistent with the findings of \citeposs{McNeil_2022}, who reported a decrease in the proportion of Arabizi comments on \textit{Tunisia-Sat.com}, an online forum, and an increase in the use of Tunisian Arabic (DA) relative to MSA. \amroo{Our study is the first to study how people's perception and usage of Arabizi change over time. However, one should not generalize the observed decline in Arabizi usage, especially since our study is relatively small-scale and is more representative of participants in the age group (24-44). For instance, \citet{wafa2024arabizi} found that 
Arabizi is widely used among Egyptian youth aged 17-23, especially among those who graduated from national and international schools.}

\paragraph{Diverse Views Are Not Necessarily Polarized} 
Our survey provided participants with a platform to express their perspectives and experiences, which may help identify common ground without labeling particular language practices, or the users themselves, as inherently ``\textit{right}'' or ``\textit{wrong}.''
We note that participants who share similar views on Arabizi often use it differently. This is particularly evident among most Algerian and Egyptian participants, who both consider Arabizi useful only in limited contexts, yet Algerians show a higher degree of reliance on Arabizi than Egyptians.
While individuals' acceptance of Arabizi varies, our qualitative analysis indicates that users and opponents of Arabizi are not necessarily in conflict.
Seemingly polarized views are often shaped by local social circumstances, which differ across communities.

\paragraph{Arabizi as a Needed Writing System for Some Arabic Speakers}
A notable proportion of participants rely on Arabizi, mainly when communicating with learners, non-native speakers, \amroo{or native speakers who can not read the Arabic script}. %
Supporting Arabizi is particularly needed for models that process user-generated content, such as content moderation systems. The dialectal variation in Arabizi spelling norms suggests that these models should be evaluated on data from speakers of different dialects to ensure their generalizability.
Lastly, as noted by \citet{keleg-2025-llm}, we need to examine Arabic speakers' reliance on Arabizi when interacting with LLMs, following similar work in other languages \cite{blaschke-etal-2024-dialect,al-kautsar-etal-2025-indonesians}.%

\todo[color=green]{[IDEA]\\
If Arabizi is hard to read, then there is minimal need to have models generating Arabizi?}

\section{Conclusion}
We present the largest human-centric cross-dialectal study of Arabizi, examining both perceptual and linguistic variation through a survey of \nparticipants participants who speak multiple Arabic dialects. Our study reveals that, although Arabizi usage patterns may be declining, particularly in Egypt, acceptance of the script varies among speakers, differences in acceptance do not necessarily correspond to divergent reasons, and Arabizi is a useful writing system for some speakers. %
To explore linguistic variation, we created \lexicon, a lexicon of \lexiconWords~Arabic words (\lexiconWordsPerDialect per dialect) with an average of 20.8 transliterations per word, along with \parallelnadi, a dataset of \nUniqueSentences parallel Arabic–Arabizi sentences.
Cross-dialectal spelling variation is evident and validated by annotators’ near-perfect identification of same-country transliterations and their general recognition of spelling cues of other dialects.

\section*{Limitations}
Our comparative study of Arabizi from both social and linguistic perspectives highlights its role as a non-secondary writing system and demonstrates its rich linguistic variation. While our work opens new research directions, it also provides opportunities for further expansion.

For \textbf{the social aspect}, our survey participants are skewed toward younger age groups, which may under-represent the perspectives of older generations. Although most participants came from five countries (Algeria, Egypt, Lebanon, Morocco, and Tunisia), reflecting the potential relevance of Arabizi in these regions as pointed out by previous work; future targeted surveys could complement our findings by exploring the views of Arabic speakers in other regions, particularly the Gulf and the broader Levant.

For \textbf{the investigation of cross-dialectal spelling variation}, we collected judgments of acceptability but did not ask participants to provide explanations. Valuable insights could be gained from a contrastive analysis of responses to parallel transliterations of the same Arabic-script sentence. By releasing our datasets, we provide the community with resources to pursue these investigations further. While we do not claim that our datasets or resources are fully representative of any dialect's usage, this work represents an important step toward understanding cross-dialectal Arabizi practices while centering participants'~perspectives.

\section*{Ethical Considerations}
This study was approved by the research ethics committee of the University of Edinburgh, School of Informatics, with reference number 813990. \amroo{We used personal connections to recruit the annotators for the parallel corpus, paying them well above the minimum wage in their countries.}

We acknowledge the potential socio-cultural biases that may be present in our data, arising from the sources or the annotation process. Our primary goal was to capture common perceptions of native speakers, and our findings are not intended to be comprehensively standardized for any specific language variation. That is, we do not claim that our datasets or resources are fully representative of any dialect's usage.
However, this work provides a step toward understanding cross-dialectal Arabizi usage while centering participants and adhering to responsible data collection practices. Finally, while we took care of manually removing instances that contain inappropriate or offensive utterances, some might have remained unnoticed.

\section*{Acknowledgments}
We thank Taha Tobaili for the early discussions on this project. We are also grateful to the generosity of our survey's participants, and to our annotators: Abderrahmane, Aziz, Fassiha, Fatima, Houda, Jana, Khalil, Moshira, Reem, and the others.

\section*{Positionality Statement}
Four authors are affiliated with governmental and academic institutions in the UAE, Lebanon, and the UK. All the authors are able to use Arabizi.
\newline
Amr Keleg is linguistically proficient in Arabic (Egyptian and MSA), and English. He has lived in Egypt, the UK, and the UAE. He frequently used Arabizi as a teenager, but currently rarely uses it.
\newline
Ahmed Amine Ben Abdallah is linguistically proficient in Arabic (Tunisian and MSA), English, and French. He has lived in Tunisia throughout his life. He has used Arabizi extensively since his teenage years and continues to use it frequently in informal online communication.
\newline
Taha Yassine is linguistically proficient in Arabic (Moroccan and MSA), English, and French. He has lived in Morocco, France, and the UK. He has always used Arabizi as his primary form of online written communication in Moroccan Arabic, using the Arabic script only when writing in MSA.
\newline
Chadi Helwe is linguistically proficient in Arabic (Lebanese and MSA), English, and French. He has lived in Lebanon, France, and Saudi Arabia. He often communicates informally online using Arabizi.
\newline
Imane Guellil is linguistically proficient in Arabic (Algerian, Tunisian, Moroccan, and MSA), English, and French. She currently resides in the UK but has previously lived in Algeria. She started using Arabizi more than 10 years ago, and still uses it in different media (e.g., WhatsApp, Facebook). 
\newline
Nedjma Ousidhoum is linguistically proficient in Arabic (Algerian, MSA, and Classical), French, and English. She has lived in Algeria, Hong Kong, and the UK. She frequently used Arabizi up to about a decade ago, but currently does not often use it; mainly when code-switching or when using a device with no easily available Arabic keyboard.

\bibliography{custom}
\newpage
\appendix
\setcounter{table}{0}
\setcounter{figure}{0}
\renewcommand{\thetable}{\Alph{section}\arabic{table}}
\renewcommand{\thefigure}{\Alph{section}\arabic{figure}}

\section{\amro{Dialects Considered in Previous Arabizi Studies}}
\label{sec:dialects_in_studies}

\amro{Different Arabic dialects were considered in Arabizi studies, yet most studies focused on a single dialect. To quantitatively estimate the dialects most represented in Arabizi studies, we queried \textit{Semantic~Scholar}'s API\footnote{\url{https://www.semanticscholar.org/product/api}} with the following keywords: \textit{Arabizi, Arabish, Arabisi, ``Franco~Arabic'', and ``Franco~Arab''}. We manually identified the dialect(s) studied in each of the 244 retrieved papers/theses. After discarding 74 papers not related to Arabizi, the top five dialects represented in both sociolinguistic studies and NLP papers/datasets were: Algeria~(42), Tunisia~(30), Morocco~(24), Egypt~(24), and Lebanon~(20).\footnote{Note that 8 papers considered Arabizi as a single variety across dialects, 4~were survey papers, 4~were duplicated entries, and 3~studied Romanized MSA. We failed to identify the varieties considered in six publications.}}

\amro{Other represented dialects are: Jordan~(16), Saudi~Arabia~(13), Palestine~(8), United~Arab~Emirates~(7), Kuwait~(5), Oman~(4), and Libya~(1). However, most of the manuscripts that consider these dialects are sociolinguistic studies, and the few that released NLP datasets were not published in *CL conferences/workshops.}

\section{Analysis of the Users' Preferred Scripts}
\label{sec:analysis_scripts}

\begin{figure*}[!tphb]
    \includegraphics[]{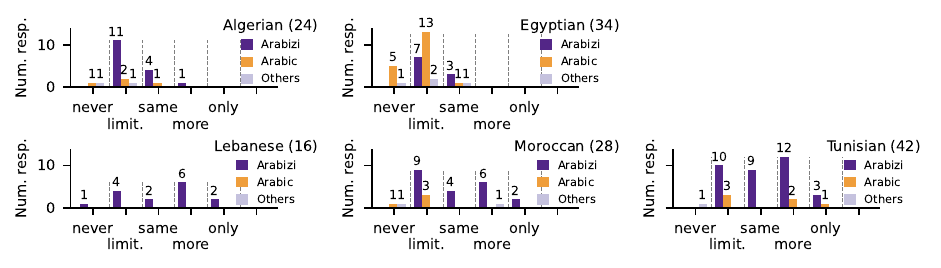}
    \caption{The distribution of the responses to \textit{(\ref{survey:response_Arabizi}) \QResponseArabizi}. The responses are grouped according to the perception patterns of \textit{(\ref{survey:perception}) \Qsurveyperception} \textbf{never}: \textit{should never be used}, \textbf{limit.}: \textit{is useful/needed in very limited contexts}, \textbf{same}: \textit{is as useful for typing as the Arabic script}, \textbf{more}: \textit{is more preferable for typing than the Arabic script}, \textbf{only}: \textit{is the only way I use for typing Arabic words}.}
    \label{fig:response_Arabizi_variation_country}
\end{figure*}

\begin{figure*}[!tbhp]
   \includegraphics[]{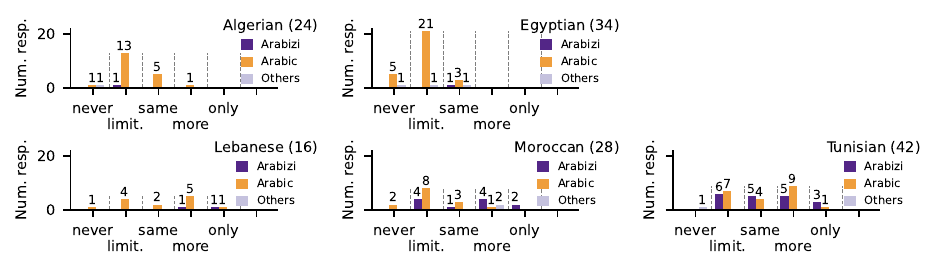}
   \caption{The distribution of the responses to \textit{(\ref{survey:response_Arabic}) \QResponseArabic}. The responses are grouped according to the perception patterns of \textit{(\ref{survey:perception}) \Qsurveyperception} \textbf{never}: \textit{should never be used}, \textbf{limit.}: \textit{is useful/needed in very limited contexts}, \textbf{same}: \textit{is as useful for typing as the Arabic script}, \textbf{more}: \textit{is more preferable for typing than the Arabic script}, \textbf{only}: \textit{is the only way I use for typing Arabic words}.}
    \label{fig:response_Arabic_variation_country}
\end{figure*}

The following questions investigate the scripts that each participant uses in the following scenarios: \textit{(\ref{survey:response_Arabizi}) \QResponseArabizi} and \textit{(\ref{survey:response_Arabic}) \QResponseArabic}

\autoref{fig:response_Arabizi_variation_country} and \autoref{fig:response_Arabic_variation_country} groups the users according to their degree of usage of Arabizi, and shows the respective scripts that they use in both scenarios. We acknowledge that multiple factors impact the users' choice of script, other than the script used by the interlocutor. Nevertheless, the distributions are indicative of clear patterns for users of different countries.

Overall, Egyptian participants show a stronger preference for using the Arabic script when replying to both Arabizi and Arabic-script messages. In contrast, Algerian and Lebanese respondents tend to match the script used by their interlocutors, while Tunisian and Moroccan respondents appear more likely to use Arabizi in both scenarios than participants from the other countries.

\section{Statistical Significance of Perception and Usage Patterns}
\label{sec:stat_sign}
\amr{To test whether the perception and usage patterns in \autoref{fig:perception_variation_country} and \autoref{fig:usage_variation_country} are different, we use Fisher's Exact Test \cite{fisher1935lady}. The test is used to determine if the difference in perception or usage distributions between a pair of countries is statistically significant or not.}

\begin{figure*}[tbph]
    \begin{subfigure}{0.48\textwidth}
        \centering
        \includegraphics[]{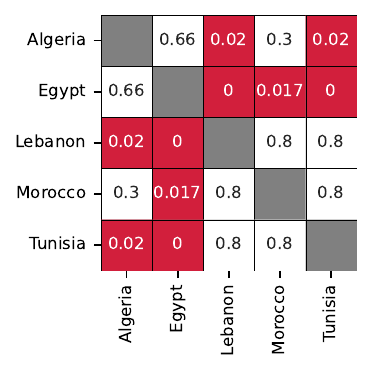}
        \caption{Perception patterns shown in \autoref{fig:perception_variation_country}.}
    \end{subfigure}
    \begin{subfigure}{0.48\textwidth}
        \centering
        \includegraphics[]{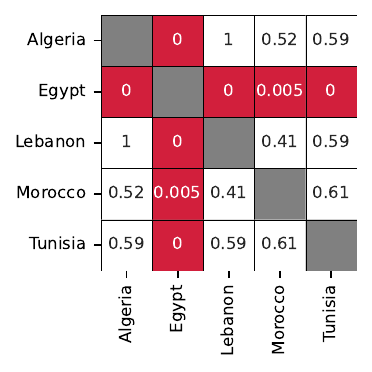}
        \caption{Usage patterns shown in \autoref{fig:usage_variation_country}.}
    \end{subfigure}
    \caption{\amr{The p-values of Fisher's Exact Tests for the different pairs of countries.} The \textcolor{red}{red} \amr{cells signify the country pairs with statistically significant (p<0.05) differences}.}
    \label{fig:stat_sign}
\end{figure*}

\amro{We use the False Discovery Rate - Benjamini/Hochberg (FDR-BH) method to adjust the computed p-values of Fisher's Tests for the ten unique pairs of countries. The adjusted p-values} are reported in \autoref{fig:stat_sign}. For the usage distributions, it is clear that they are clustered into two groups: (Algeria, Egypt) and (Lebanon, Morocco, Tunisia), as previously mentioned \S\ref{sec:cross_dialectal_percep}, \amr{based on the distributions shown in \autoref{fig:perception_variation_country}. Moreover, it is clear that the usage distribution of Arabizi in Egypt is statistically different from the distributions of the four other countries.}

\section{Perception and Usage Distributions for Five Other Countries}
\begin{figure*}[tbph]
    \centering
     \begin{subfigure}[b]{0.95\textwidth} 
    \includegraphics[width=\textwidth]{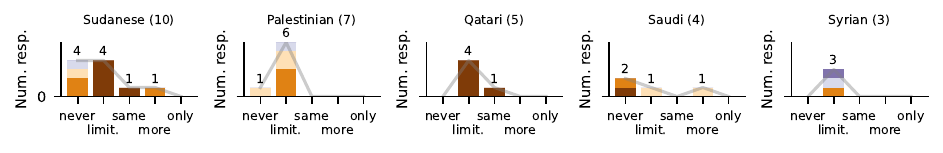}
     \caption{The distribution of the responses to \textit{(\ref{survey:perception}) \Qsurveyperception} \textbf{never}: \textit{should never be used}, \textbf{limit.}: \textit{is useful/needed in very limited contexts}, \textbf{same}: \textit{is as useful for typing as the Arabic script}, \textbf{more}: \textit{is more preferable for typing than the Arabic script}, \textbf{only}: \textit{is the only way I use for typing Arabic words}. \textbf{Note:}~Color encodes different age groups as follows:~\captionBox{a}{(<18)-24} \captionBox{b}{25-34} \captionBox{c}{35-44} \captionBox{d}{45-54} \captionBox{e}{55+}}
     \label{fig:perception_variation_country_others}
     \end{subfigure}

     \begin{subfigure}[b]{0.95\textwidth} 
    \includegraphics[width=\textwidth]{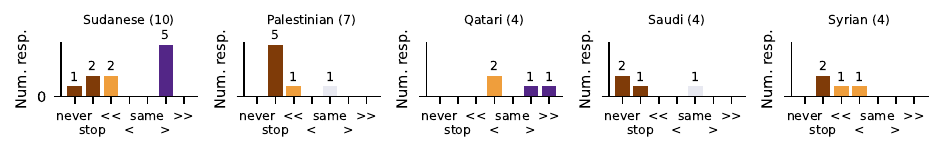}
    \caption{The distribution of the responses to \textit{(\ref{survey:usage}) \QUsage} \textbf{never}: answered \textit{no} to \textit{(\ref{survey:ever_used}) \QEverUsedArabizi}, \textbf{stop}: \textit{I stopped using Arabizi}, \textbf{<\hspace{0pt}<}: \textit{Much less now compared to when I started texting}, \textbf{<}: \textit{Somewhat less now compared to when I started texting}, \textbf{same}: \textit{About the same}, \textbf{>}: \textit{Somewhat more now compared to when I started texting}, \textbf{>\hspace{0pt}>}: \textit{Much more now compared to when I started texting}. \textbf{Note}:~Color encodes choices having the same direction (e.g, increase/decrease).}
    \label{fig:usage_variation_country_others}
    \end{subfigure}
   \caption{The distributions of the perception (\ref{fig:perception_variation_country_others}) and usage of Arabizi (\ref{fig:usage_variation_country_others}) in five additional countries.}
    \label{fig:response_Arabic_variation_country_others}

\end{figure*}

\amr{For completeness, we report the distributions of the perception and usage patterns for the responses of the participants from Sudan, Palestine, Qatar, Saudi Arabia, and Syria in \autoref{fig:response_Arabic_variation_country_others}.}

\section{Aligning Arabic and Latin Letters in Arabic/Arabizi Word Pairs}

\input{algorithm}

\input{letter_alignment}

Starting from the lexicon of Arabic words with their corresponding transliterations provided by the survey participants, we used the algorithm shown in  \autoref{lst:align_word} to align the Arabic words with their transliterations. An Arabic letter is aligned and consumed only if the corresponding Arabizi letter(s) belong(s) to the mappings specified in the table. Vowels in Latin script may also align with an empty string in Arabic script to consider short vowels (i.e., diacritics) commonly dropped in Arabic. The algorithm \amr{(listed in \autoref{lst:align_word})} recursively searches for the alignment that maximizes the number of Latin letters mapped to Arabic ones, under the constraint that no Arabic consonant may be skipped. We progressively analyzed the words that could not be aligned and manually identified new letter mappings not included in the initial seed table (\textbf{in bold} in \autoref{table:sub_letter_mappings_transposed}).

Before applying the alignment algorithm, we applied some preprocessing steps to the collected transliterations. 248 out of 4,718 total transliterations had typos and were discarded (e.g., transliterating \AR{ماء} as \textit{lma} instead of just \textit{ma}). Moreover, only \amro{31} entries had multiple possible transliterations for a single word, which we treat as independent transliterations for the word.

Afterward, we aggregated the Latin equivalents of each Arabic letter based on the alignments of the Arabic words and their transliterations. The aggregated counts are shown in \autoref{table:letter_mappings}.

\subsection{Additional Inter/Intra-Dialectal Spelling Variation Patterns}
\label{sec:spelling_variation_extended}

\begin{itemize}[leftmargin=*, noitemsep, nolistsep]
    \item \textit{i} is the most common transliteration of \AR{ي} in all the dialects except for Egyptian.
    \item \AR{ث} /$\theta$/ is mapped to \textit{t} (the letter expected for \AR{ت}) almost categorically for Moroccan Arabizi, with lesser frequencies for Egyptian and Algerian Arabizi.
    \item \AR{ء} is commonly dropped in Arabizi for Algeria and Tunisia.
    \item \textit{3} and \textit{7} are consistently used to encode \AR{ع} and \AR{ح} respectively. \textit{2} is typically used for \AR{ء} in the 5 dialects, but also for \AR{ق} in Egypt and Lebanon. \textit{4} is rarely used for \AR{ض}, \AR{ذ}, \AR{ظ} in Morocco and for \AR{ش} in Egypt. \textit{9} is only used in Algeria, Morocco, and Tunisia to refer to \AR{ق} and to \textkurdish{ڤ} in Algeria and Morocco.
    \item As previously noted in the literature, \textit{ch} (a sign of being francophone) is used in all countries but Egypt for rendering the letter \AR{ش}. \todo{check the impact of the second language that each participant has chosen on their mapping of Arabic letters into Arabizi. For instance, are they consistent?}
\end{itemize}

\taha{Discuss how some letters in particular settings are handled. E.g. \AR{ه} or \AR{ة} at the end of feminine words usually transliterated to "a" (at least in Moroccan Darija).}

\section{Lebanese Labels}
\label{sec:lebanese_annotations}

Instead of having the full MLADI dataset labeled by Lebanese annotators, we identified samples valid in at least one of the other four considered dialects, with respective country-level ALDi scores that are more than 0.33. These samples include ones that are valid in at least one country and potentially valid in Lebanon, or others that are valid in at least two countries and potentially valid in Lebanon. In total, 390 samples were identified and labeled by three Lebanese annotators, using the same guidelines in \citet{abdul-mageed-etal-2024-nadi}.

\todo{Talk about the interannotator agreement, and the stats of this dataset}

\
\section{Arabizi Sentence-level Annotation}

\subsection{Additional Patterns Identified for the Arabizi Sentence Annotation}

\begin{itemize}[leftmargin=*, noitemsep, nolistsep]
\item The 46 sentences common to Algeria and Morocco revealed a striking \textbf{opposite pattern}: both Algerian annotators (A, B) labeled most Moroccan transliterations as \textit{My Country}, whereas annotator Morocco~(G) correctly identified Algerian transliterations, and Morocco~(H) did not.
\item Lebanese annotators labeled more Egyptian transliterations as \textit{My Country} than Egyptian annotators did for Lebanese transliterations.
\end{itemize}

\subsection{Annotation Interface for Judging Transliterations}
\begin{figure*}[!h]
    \centering
    \includegraphics[width=0.9\linewidth]{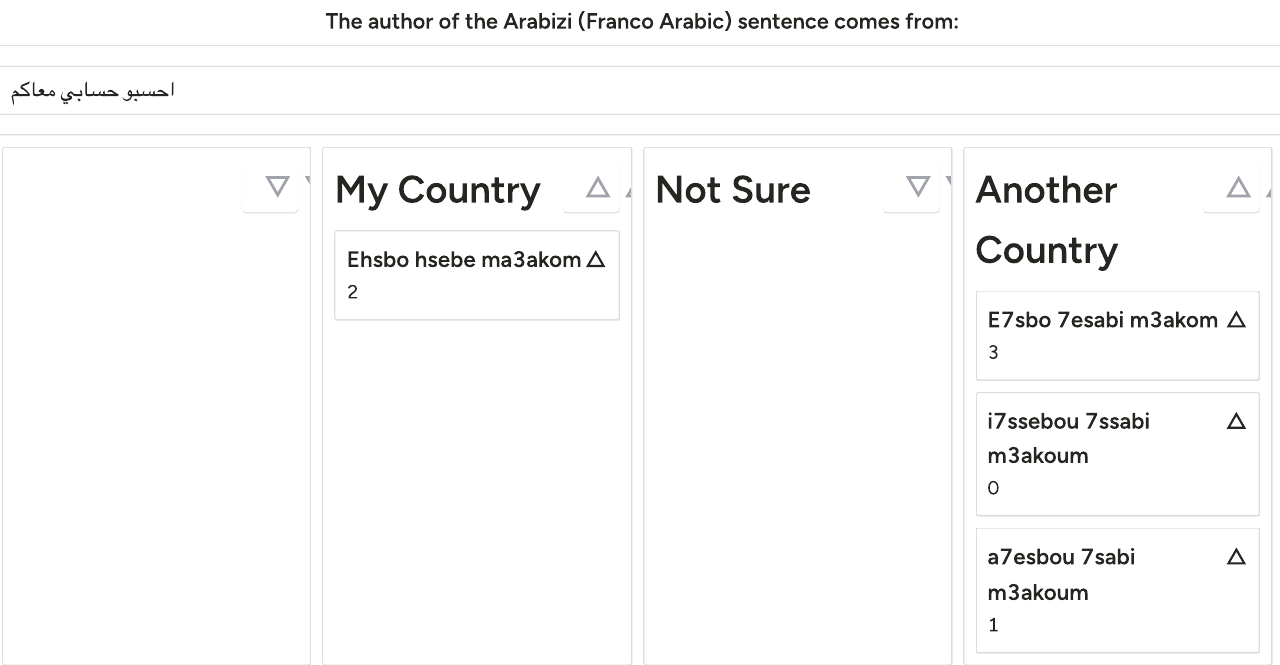}
    \caption{A set of four transliterations annotated by one of the Lebanese participants.}
    \label{fig:annotation_interface}
\end{figure*}

For the experiment described in \S\ref{sec:experiment}, we used LabelStudio to present the transliteration groups. A screenshot of the interface is shown in \autoref{fig:annotation_interface}.

\section{Survey's Questions}
\label{sec:survey_questions}

\taha{TODO - add the questions in Arabic, English, and French + indicate the ones that are optional and the ones that are only shown based on answers to preceding questions.}

This appendix reproduces the complete structure of the questionnaire as presented to participants. All texts were available in three languages.

\noindent\textbf{Language legend:} \langEN{English} \quad \langFR{Français} \quad \langAR{العربية}

\subsection{Global Instructions}
\langEN{This survey investigates the variations in writing Arabic characters using Latin letters and numbers, a practice often called Arabizi (sometimes referred to as Arabish or Franco-Arabic).\\
For example, writing \AR{"مرحبا"} as "marhaba" or "mar7aba".}

\langFR{Cette enquête étudie les variations dans l’écriture des caractères arabes à l’aide de lettres et de chiffres latins, une pratique souvent appelée Arabizi (parfois désignée sous le nom d'Arabish ou Franco-Arabe).\\
Par exemple, écrire \AR{"مرحبا"} comme "marhaba" ou "mar7aba".}

\langAR{يبحث هذا الاستطلاع في الاختلافات في كتابة الكلمات العربية باستخدام الحروف والأرقام اللاتينية، وهي طريقة تعرف غالبًا باسم الأرابيزي (يشار إليها أحيانًا باسم أرابيش أو فرانكو أراب).\\
على سبيل المثال، كتابة "مرحبا" كـ "marhaba" أو "mar7aba".}

~\\~\\
The survey is split into 2 parts:
\begin{itemize}
    \item \textbf{Part 1} -- A quick questionnaire in which you will be asked about how you use and perceive Arabizi.
    \item \textbf{Part 2} -- Follow-up questions for how you write words in Arabizi.
\end{itemize}

\langFR{
Le questionnaire est divisé en 2 parties :
\begin{itemize}
    \item \textbf{Partie 1} -- Un questionnaire rapide sur votre utilisation et perception de l'Arabizi.
    \item \textbf{Partie 2} -- Des questions complémentaires sur votre manière d'écrire certains mots en Arabizi.
\end{itemize}
}

\langAR{
تم تقسيم الاستطلاع إلى قسمين:
\begin{itemize}
    \item \textbf{الجزء الأول} --استبيان سريع يُطلب فيه منك تحديد كيفية استخدامك ورأيك في الأرابيزي.
    \item \textbf{الجزء الثاني} --دراسة متابعة لدراسة كيفية كتابة بعض الكلمات بالأرابيزي.
\end{itemize}
}

\newpage

\subsection{Part 1 -- Background, Usage, and Perception}
\label{sec:survey_part1_detailed_questions}

\begin{figure*}[hptb]
    \centering
    \includegraphics[width=0.9\linewidth]{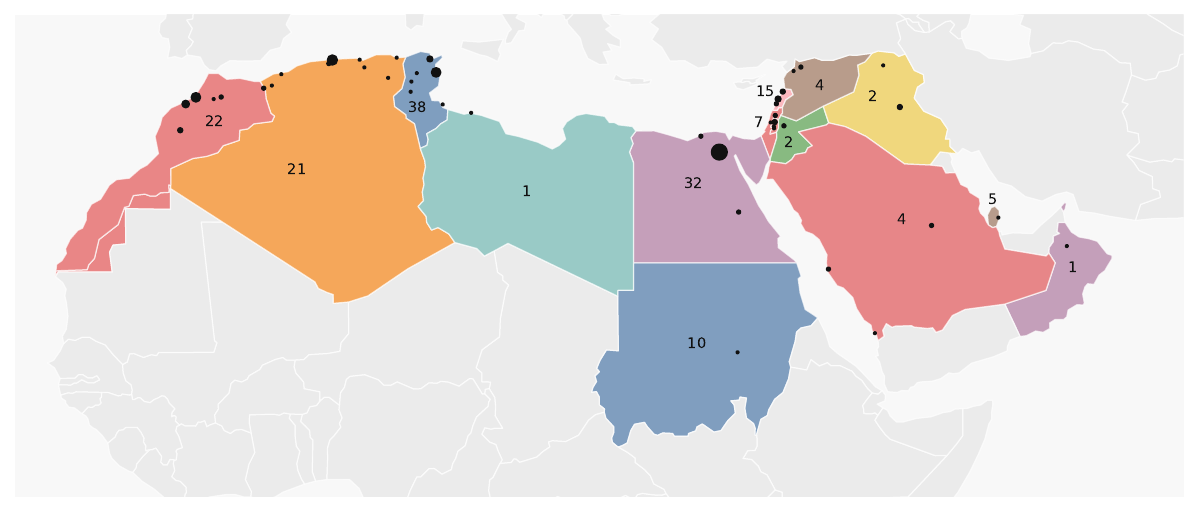}
    \caption{Geographic distribution of survey participants based on responses to \textit{(\ref{survey:response_country})} and \textit{(\ref{survey:response_city})}. Each country is labeled with the number of participants, and the size of the black dots is proportional to the number of respondents from each city. \textbf{Note:} Participants who did not specify their location or provided incomplete information are not shown.}
    \label{fig:map}
\end{figure*}

In this part, participants were asked about their background, use, and perception of Arabizi.

\begin{enumerate}[label=\textbf{Q\arabic*}, leftmargin=1cm]
\item \label{survey:response_country}
\langEN{What is your native country-level Arabic dialect?} \\
\langFR{Quel est votre dialecte arabe maternel ?} \\
\langAR{ما هي لهجتك العربية؟}

\begin{itemize}[label=$\square$]
    \item Algerian
    \item Egyptian
    \item Lebanese
    \item Moroccan
    \item Tunisian
    \item Other (Please specify): \_\_\_
\end{itemize}

\begin{figure}[H]
    \centering
    \includegraphics{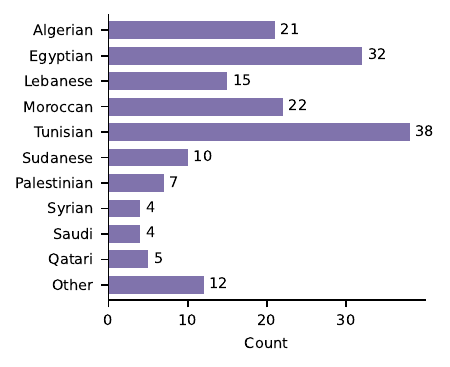}
\end{figure}

\item \langEN{What is your second language?} \\
\langFR{Quelle est votre deuxième langue ?} \\
\langAR{ما هي لغتك الثانية؟}

\begin{itemize}[label=$\square$]
    \item English
    \item French
    \item Spanish
    \item Italian
    \item Other (Please specify): \_\_\_
\end{itemize}

\begin{figure}[H]
    \centering
    \includegraphics{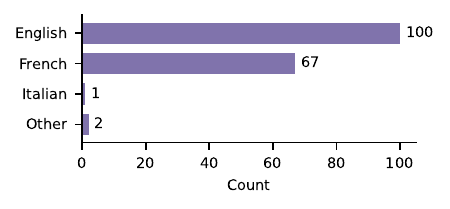}
\end{figure}

\newpage

\item \label{survey:response_city}
\langEN{Which city/province's dialect do you speak?\\
For example: Cairenese or Sa'idi in Egypt, Rabat's dialect in Morocco, ...} \\
\langFR{Quelle est la ville ou la province d'origine du dialecte que vous parlez ?\\
Par exemple : le cairote ou le sa'idi en Égypte, le dialecte de Rabat au Maroc, ...} \\
\langAR{من أيّ مدينة أو محافظة لهجتك؟\\
علي سبيل المثال: اللهجة القاهرية أو الصعيدية في مصر واللهجة الرباطية في المغرب ...}

\textit{Optional. Free text.}

\item \langEN{Have you been currently living abroad for longer than 6 consecutive months?} \\
\langFR{Vivez-vous actuellement à l'étranger? (Veuillez répondre oui si cela est le cas depuis plus de 6 mois consécutifs.)} \\
\langAR{هل تقيم حاليًا في الخارج (لمدة تزيد عن ستة أشهر متتالية)؟}

\textit{Optional}

\begin{itemize}[label=$\square$]
    \item Yes
    \item No
\end{itemize}

\begin{figure}[H]
    \centering
    \includegraphics{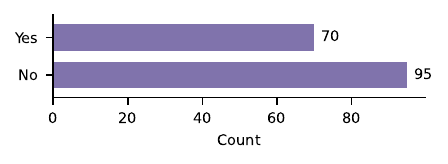}
\end{figure}

\item \langEN{If you have been currently living abroad for longer than 6 consecutive months, please specify if you are living in another Arab country or a non-Arab one.} \\
\langFR{Si vous vivez actuellement à l'étranger depuis plus de 6 mois consécutifs, veuillez préciser si vous vivez dans un autre pays arabe ou dans un pays non arabe.} \\
\langAR{في حالة إقامتك حاليًا في الخارج لمدة تزيد عن ستة أشهر متتالية برجاء تحديد إذا كنت تقيم في دولة عربية أخرى أو في دولة غير عربية.}

\textit{Optional. Free text. Only asked if the answer to Q4 was Yes.}

\begin{itemize}[label=$\square$]
    \item An Arab country (please specify if you want): \_\_\_
    \item A non-Arab country (please specify if you want): \_\_\_
\end{itemize}

\begin{figure}[H]
    \centering
    \includegraphics{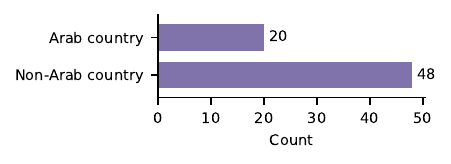}
\end{figure}

\item \langEN{What is your age group?} \\
\langFR{Quelle est votre tranche d'âge ?}\\
\langAR{ما هي فئتك العمرية؟}

\begin{itemize}[label=$\square$]
    \item Under 18
    \item 18--24
    \item 25--34
    \item 35--44
    \item 45--54
    \item 55--64
    \item 65+
    \item Prefer not to say
\end{itemize}

\begin{figure}[H]
    \centering
    \includegraphics{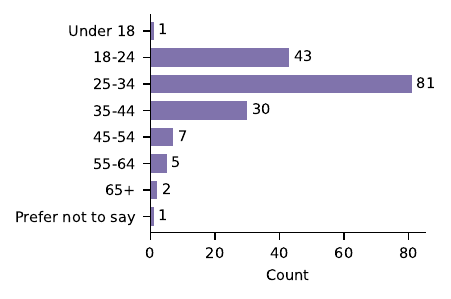}
\end{figure}

\item \label{survey:perception} \Qsurveyperception \\
\langFR{Pour vous, l'Arabizi ...} \\
\langAR{بالنسبة إليك، نظام الأرابيزي ...}

\begin{itemize}[label=$\square$]
    \item should never be used.
    \item is useful/needed in very limited contexts.
    \item is as useful for typing as the Arabic script.
    \item is more preferable for typing than the Arabic script.
    \item is the only way I use for typing Arabic words.
    \item Others (please specify): \_\_\_
\end{itemize}

\begin{figure}[H]
    \centering
    \includegraphics{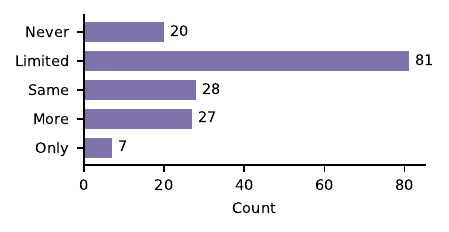}
\end{figure}

\item \label{survey:why_perception} \QsurveyWhyPerception \\
\langFR{Veuillez expliquer pourquoi vous pensez cela.} \\
\langAR{يرجى توضيح سبب اعتقادك ذلك.}

\textit{Optional. Free text.}

\newpage

\item \label{survey:ever_used} \QEverUsedArabizi \\
\langFR{Avez-vous déjà utilisé l'Arabizi ?} \\
\langAR{هل سبق لك  استخدام الأرابيزي؟}

\begin{itemize}[label=$\square$]
    \item Yes
    \item No
\end{itemize}

\begin{figure}[H]
    \centering
    \includegraphics{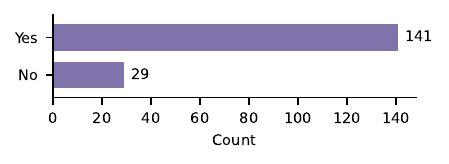}
\end{figure}

\item \langEN{When did you start texting on mobile phones? Please identify a specific year (YYYY).} \\
\langFR{Quand est-ce que vous avez commencé à envoyer des SMS sur un téléphone mobile ?\\
Veuillez indiquer une année spécifique (AAAA).} \\
\langAR{متى بدأت إرسال الرسائل النصية عبر الهواتف المحمولة؟\\
يرجى تحديد سنة معينة (YYYY).}

\textit{Free text. Only asked if the answer to Q9 was Yes.}

\begin{figure}[H]
    \centering
    \includegraphics{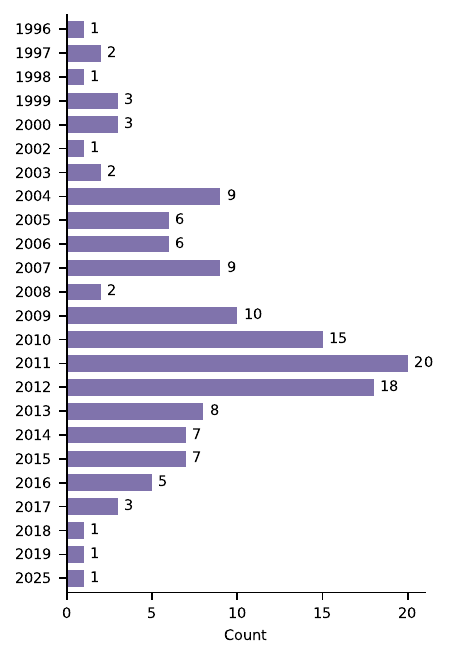}
\end{figure}

\newpage

\item \label{survey:usage} \QUsage \\
\langFR{Comment votre usage de l'Arabizi a-t-il évolué depuis que vous avez commencé à envoyer des SMS ?} \\
\langAR{ما هي نسبة استخدامك للأرابيزي الآن مقارنة بأعلى نسبة استخدام منذ بدأت إرسال الرسائل النصية عبر الهواتف المحمولة؟}

\begin{itemize}[label=$\square$]
    \item I stopped using Arabizi
    \item Much less now compared to when I started texting
    \item Somewhat less now compared to when I started texting
    \item About the same
    \item Somewhat more now compared to when I started texting
    \item Much more now compared to when I started texting
\end{itemize}

\begin{figure}[H]
    \centering
    \includegraphics{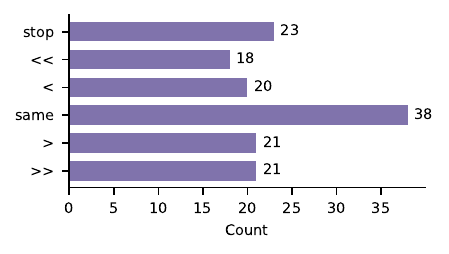}
\end{figure}

\item \langEN{In what context do you use Arabizi? (Please select all that applies)} \\
\langFR{Dans quel contexte utilisez-vous l'Arabizi ? (Veuillez sélectionner toutes les réponses qui s'appliquent)} \\
\langAR{في أي سياق تستخدم الأرابيزي؟ (يُرجى اختيار كل ما ينطبق)}

\textit{Only asked if the answer to Q11 wasn't "I stopped using Arabizi".}

\begin{itemize}[label=$\square$]
    \item Texting/chatting
    \item Social Media
    \item Others (please specify): \_\_\_
\end{itemize}

\begin{figure}[H]
    \centering
    \includegraphics{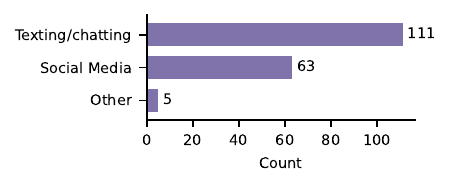}
\end{figure}

\item \langEN{Which social media platform do you usually use?} \\
\langFR{Quels sont les réseaux sociaux que vous utilisez ?} \\
\langAR{ما هي منصة التواصل الاجتماعي التي تستخدمها عادة؟}

\begin{itemize}[label=$\square$]
    \item X (formerly Twitter)
    \item Facebook
    \item YouTube
    \item TikTok
    \item Instagram
    \item Reddit
    \item Telegram
    \item WhatsApp
    \item Other (please specify): \_\_\_
\end{itemize}

\begin{figure}[H]
    \centering
    \includegraphics{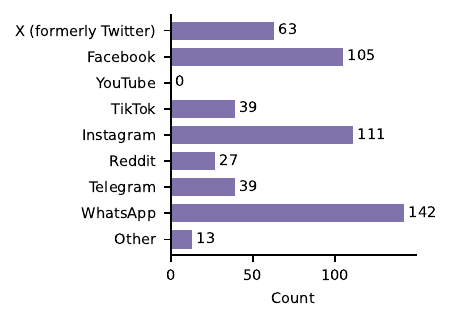}
\end{figure}

\item \langEN{Which social media platform do you think people tend to use Arabizi on?} \\
\langFR{Quels sont les réseaux sociaux sur lesquels vous trouvez que les gens ont tendance à utiliser l'Arabizi ?} \\
\langAR{ما هي منصة التواصل الاجتماعي التي تعتقد أن الناس يميلون إلى استخدام الأرابيزي عليها؟}

\begin{itemize}[label=$\square$]
    \item X (formerly Twitter)
    \item Facebook
    \item YouTube
    \item TikTok
    \item Instagram
    \item Reddit
    \item Telegram
    \item WhatsApp
    \item Other (please specify): \_\_\_
\end{itemize}

\begin{figure}[H]
    \centering
    \includegraphics{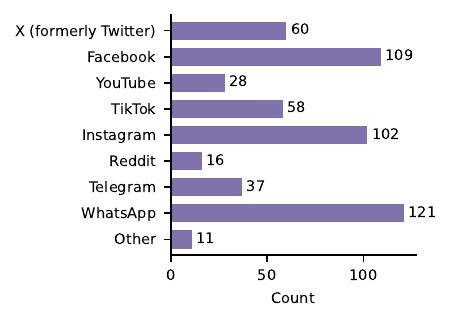}
\end{figure}

\item \label{survey:response_Arabizi} \QResponseArabizi \\
\langFR{Si quelqu'un vous écrit en Arabizi, êtes-vous plus susceptible de répondre en Arabizi ou en utilisant l'alphabet arabe ?} \\
\langAR{إذا وصلتك رسالة بالأرابيزي، هل من المرجح أن ترد بالأرابيزي أم باستخدام الحروف العربية؟}

\begin{itemize}[label=$\square$]
    \item Arabic script
    \item Arabizi
    \item Others (Please specify): \_\_\_
\end{itemize}

\begin{figure}[H]
    \centering
    \includegraphics{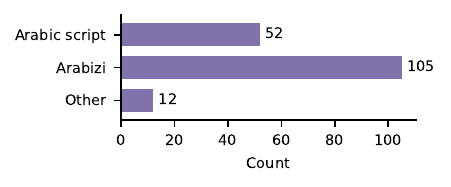}
\end{figure}

\newpage

\item \label{survey:response_Arabic} \QResponseArabic \\
\langFR{Si quelqu'un vous écrit en utilisant l'alphabet arabe, êtes-vous plus susceptible de répondre en Arabizi ou en utilisant l'alphabet arabe ?} \\
\langAR{إذا وصلتك رسالة بالحروف العربية، هل من المرجح أن ترد بالأرابيزي أم باستخدام الحروف العربية؟}

\begin{itemize}[label=$\square$]
    \item Arabic script
    \item Arabizi
    \item Others (Please specify): \_\_\_
\end{itemize}

\begin{figure}[H]
    \centering
    \includegraphics{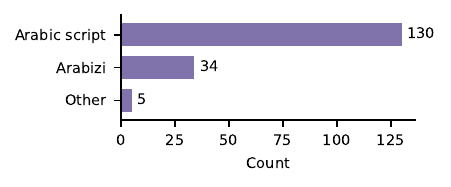}
\end{figure}

\item \label{survey:comments} \QsurveyComments \\
\langFR{Avez-vous des commentaires à partager avant de passer à la deuxième partie ?} \\
\langAR{هل هناك أي تعليقات تريد مشاركتها معنا قبل بدء الجزء الثاني؟}

\textit{Optional. Free text.}

\end{enumerate}

\subsection{Part 2 -- Rendering Arabic Words in Arabizi}
In this part, participants were asked how they would write several Arabic words in Arabizi. The following prompt was shown above each item:
~\\

\noindent\langEN{How do you write the word \textbf{[word]} in Arabizi?}

\noindent\langFR{Comment écrivez-vous le mot \textbf{[mot]} en Arabizi ?}

\noindent\langAR{ كيف تكتب كلمة \textbf{[كلمة]} بالأرابيزي؟}

\begin{table*}[th!]
\scriptsize
\caption{Part 2 word list grouped by dialect (1/2): \textbf{Algerian}, \textbf{Egyptian}, \textbf{Lebanese}. Each dialect has the Arabic item and its English and French glosses.}
\label{tab:part2_words_by_dialect_a}
\centering
\begingroup
\setlength{\tabcolsep}{4pt}
\resizebox{\textwidth}{!}{%
\begin{tabular}{lll lll lll}
\toprule
\multicolumn{3}{c}{\textbf{Algerian}} & \multicolumn{3}{c}{\textbf{Egyptian}} & \multicolumn{3}{c}{\textbf{Lebanese}} \\
\cmidrule(lr){1-3} \cmidrule(lr){4-6} \cmidrule(lr){7-9}
\textbf{Arabic} & \textbf{English} & \textbf{French} & \textbf{Arabic} & \textbf{English} & \textbf{French} & \textbf{Arabic} & \textbf{English} & \textbf{French} \\
\midrule
\AR{أنا} & me & moi & \AR{أنا} & me & moi & \AR{أمل} & hope & espoir\\
\AR{إنسان} & individual & individuel & \AR{إنتماء} & belonging & appartenance & \AR{إسم} & name & nom\\
\AR{آتمنا} & I hope & espoirs & \AR{آمال} & hopes & espoirs & \AR{إيام} & days & jours\\
\AR{ماء} & water & eau & \AR{جزء} & part & partie & \AR{آخر} & last & dernier\\
\AR{ماء} & water & eau & \AR{لئيم} & mean & méchant & \AR{جزء} & part & partie\\
\AR{سؤال} & question & question & \AR{سؤال} & question & question & \AR{رئيس} & president & président\\
\AR{سلام} & peace & paix & \AR{استعمال} & usage & usage & \AR{سؤال} & question & question\\
\AR{برا} & outside & dehors & \AR{استعملوا} & they used & ils ont utilisé & \AR{استعمال} & usage & utilisation\\
\AR{ماكلة} & food & nourriture & \AR{سلام} & peace & paix & \AR{سلام} & peace & paix\\
\AR{حمى} & fever & fièvre & \AR{حمى} & fever & fièvre & \AR{استعملوا} & they used & ils ont utilisé\\
\AR{بزاف} & many/a lot & beaucoup & \AR{باب} & door & porte & \AR{فتوى} & Religious ruling or decree & Règle ou décret religieux\\
\AR{بلاكار} &  &  & \AR{باركيه} &  &  & \AR{باب} & door & porte\\
\AR{مثال} & example & exemple & \AR{ثانوية} & secondary & secondaire & \AR{بيتزا} &  & \\
\AR{جامع} & mosque & mosquée & \AR{ثعبان} & snake & serpent & \AR{ثواني} & seconds & secondes\\
\AR{حاجة} & thing & chose & \AR{جمال} & beauty & beauté & \AR{ثورة} & revolution & révolution\\
\AR{صباح} & morning & matin & \AR{جراج} &  &  & \AR{دجاج} & mosque & mosquée\\
\AR{خوك} & your brother & ton frère & \AR{حاجة} & something & quelque chose & \AR{جراج} &  & \\
\AR{هذاك} & that one & celui-là & \AR{صباح} & morning & matin & \AR{حاج} & enough & assez\\
\AR{شجرة} & tree & arbre & \AR{خمسة} & five & cinq & \AR{صباح} & morning & matin\\
\AR{رصاصة} & bullet & balle & \AR{ذئب} & wolf & loup & \AR{خمسة} & five & cinq\\
\AR{ضحك} & laugh & rire & \AR{شمس} & sun & soleil & \AR{ذكي} & smart & intelligent\\
\AR{طبسي} & plate & assiette & \AR{صاحبي} & my friend & mon ami & \AR{شاورما} &  & \\
\AR{ظلمة} & darkness & obscurité & \AR{ضحك} & laugh & rire & \AR{صحتين} & Cheers & Acclamations\\
\AR{غدوة} & tomorrow & demain & \AR{طيران} & flying & vol & \AR{مضطر} & obliged & obligé\\
\AR{فرحة} & joy & joie & \AR{ظلم} & injustice & injustice & \AR{طاولة} & Table & Tableau\\
\AR{كرافاطة} & necktie & cravate & \AR{غياب} & absence & absence & \AR{مظاهرة} & Protest & manifestation\\
\AR{رقيق} & thin & mince & \AR{فراولة} & strawberry & fraise & \AR{غريب} & Strange & Étrange\\
\AR{قصبر} & coriander & coriandre & \AR{فارندا} &  &  & \AR{فلافل} &  & \\
\AR{كوشة} & oven & four & \AR{قانون} & law & loi & \AR{رقيق} & delicate & délicat\\
\AR{هرب} & to escape & s'échapper & \AR{بتقول} & (she) says & (elle) dit & \AR{قديم} & old & vieux\\
\AR{شابة‎} & beautiful & beau & \AR{كتاب} & book & livre & \AR{بكير} & early & tôt\\
\AR{جنينة الحومة} & neighbourhood park & parc de quartier & \AR{هايل} & brilliant & brillant & \AR{هوا} & Air & Air\\
\AR{وحدي} & alone & seul & \AR{جنينة} & garden & jardin & \AR{شمعة} & candle & bougie\\
\AR{عوام} & swimmer & nageur & \AR{جنينة الورد} & flowers garden & jardin de fleurs & \AR{حديقة الورد} & flowers garden & jardin de fleurs\\
\AR{روح} & go & va & \AR{ألوان} & colors & couleurs & \AR{ألوان} & colors & couleurs\\
\AR{كولو} & eat it & mange-le & \AR{معروف} & favor & service & \AR{وحدي} & alone & seul\\
\AR{يقرا} & he studies & il étudie & \AR{نمو} & growth & croissance & \AR{معروف} & Favor & Service\\
\AR{يستنى} & he waits & il attend & \AR{وادي} & valley & vallée & \AR{جدو} & my grandfather & mon grand-père\\
\AR{يسقسي} & he asks & il demande & \AR{أبو ...} & ...'s dad & ...le père de ... & \AR{حلو} & nice & bon\\
\AR{ياسمين} & jasmine & jasmin & \AR{ياسمين} & jasmine & jasmin & \AR{ياسمين} & jasmine & jasmin\\
\AR{جراچ‎} &  &  & \AR{جراچ‎} &  &  & \AR{ڤيزا} &  &  \\
\AR{تڜبتڜاق} & dish & plat & \AR{ڤارندا} &  &  &  &  &  \\
\AR{ڜكڜوكة} &  &  &  &  &  &  &  &  \\
\AR{ڤليل} & poor & pauvre &  &  &  &  &  &  \\
\AR{ڤايلة} & early afternoon & début d'après-midi &  &  &  &  &  &  \\
\AR{زفيطي} &  &  &  &  &  &  &  &  \\
\bottomrule
\end{tabular}%
}
\endgroup
\end{table*}

\clearpage

\begin{table*}[th!]
\scriptsize
\caption{Part 2 word list grouped by dialect (2/2): \textbf{Moroccan}, \textbf{Tunisian}. Each dialect has the Arabic item and its English and French glosses.}
\label{tab:part2_words_by_dialect_b}
\centering
\begin{tabular}{lll lll}
\toprule
\multicolumn{3}{c}{\textbf{Moroccan}} & \multicolumn{3}{c}{\textbf{Tunisian}} \\
\cmidrule(lr){1-3} \cmidrule(lr){4-6}
\textbf{Arabic} & \textbf{English} & \textbf{French} & \textbf{Arabic} & \textbf{English} & \textbf{French} \\
\midrule
\AR{أسئلة} & questions & questions & \AR{أنا} & me & moi\\
\AR{إنسان} & individual & individuel & \AR{إستنا} & wait & attends\\
\AR{آخر} & last & dernier & \AR{آخر} & last & dernier\\
\AR{جزء} & part & partie & \AR{ماء} & water & eau\\
\AR{رئيس} & president & président & \AR{رئيس} & president & président\\
\AR{سؤال} & question & question & \AR{سؤال} & question & question\\
\AR{استعمال} & usage & utilisation & \AR{استعمال} & usage & utilisation\\
\AR{استعملوا} & they used & ils ont utilisé & \AR{يلعبوا} & they are playing & ils jouent\\
\AR{تلفازا} & TV & TV & \AR{طيار} & pilot & pilote\\
\AR{فتوى} & Religious ruling or decree & Règle ou décret religieux & \AR{حمى} & fever & fièvre\\
\AR{باب} & door & porte & \AR{باب} & door & porte\\
\AR{بلاكار} &  &  & \AR{بلاكار} &  & \\
\AR{مثال} & example & exemple & \AR{ثما} & there is & il y a\\
\AR{شرجم} & window & fenêtre & \AR{جامع} & mosque & mosquée\\
\AR{حمام} & bathroom & salle de bain & \AR{حاجة} & thing & chose\\
\AR{صباح} & morning & matin & \AR{صباح} & morning & matin\\
\AR{سخون} & hot & chaud & \AR{خوك} & your brother & ton frère\\
\AR{مذاق} & taste & goût & \AR{هذاكا} & that one & celui-là\\
\AR{مشكيل} & problem & problème & \AR{برشا} & a lot & beaucoup\\
\AR{خلص} & to pay & payer & \AR{صاحبي} & my friend & mon ami\\
\AR{عضلة} & muscle & muscle & \AR{ضحك} & laugh & rire\\
\AR{طيارة} & airplane & avion & \AR{يعطي} & gives & donne\\
\AR{ظلم} & injustice & injustice & \AR{يظهرلي} & it appears to me & il me semble\\
\AR{غيام} & clouds & nuages & \AR{غدوة} & tomorrow & demain\\
\AR{فران} & oven & four & \AR{فلوس} & money & argent\\
\AR{كرافاطا} &  &  & \AR{فيراندا} &  & \\
\AR{رقيق} & thin & mince & \AR{قديم} & old & vieux\\
\AR{قال} & said & dit & \AR{كتاب} & book & livre\\
\AR{بكري} & early & tôt & \AR{هكا} & like this & comme ça\\
\AR{كركاع} & walnut & noix & \AR{باهية} & good & bien\\
\AR{هرب} & to escape & s'échapper & \AR{جنينة الحومة} & neighbourhood park & parc de quartier\\
\AR{هدية} & gift & cadeau & \AR{ألوان} & colors & couleurs\\
\AR{هدية العيد} & Eid gift & Cadeau de l'Aïd & \AR{معروف} & known & connu\\
\AR{ألوان} & colors & couleurs & \AR{وحدي} & alone & seul\\
\AR{روز} & rice & riz & \AR{بو} & father & père\\
\AR{وردة} & flower & fleur & \AR{ولدي} & my son & mon fils\\
\AR{يشربو} & they drink & ils boivent & \AR{ياسمين} & jasmine & jasmin\\
\AR{بياع} & shopkeeper & commerçant & \AR{يمشي} & my love & mon amour\\
\AR{خالي} & my uncle & mon oncle & \AR{يڤدم} & bite & mordre\\
\AR{رقيق} & thin & mince & \AR{ڥيلا} &  &  \\
\AR{ياسمين} & jasmine & jasmin &  &  &  \\
\AR{ڤيلا} &  &  &  &  &  \\
\AR{ڭيطار} &  &  &  &  &  \\
 &  &  &  &  &  \\
 &  &  &  &  &  \\
\bottomrule
\end{tabular}
\end{table*}

\clearpage

\onecolumn

\section{\lexicon's Words and Transliterations}
\label{sec:lexicon_freq}
This section reports, for each dialect, all Arabizi transliterations collected in our survey, along with their frequencies, including those we manually identified as incorrect. For every Arabic word, the right column lists all distinct transliterations observed, with counts in parentheses indicating how many responses used each form. Within each row, transliterations are ordered by decreasing frequency. Signs of regularity---indicated by the presence of a general preference for a specific form---and sparsity---indicated by a different number of forms for each word---can be noticed, providing more evidence of both aspects (refer to \S\ref{sec:background}).

\subsection{Algerian words}
\begin{longtable}{p{0.08\textwidth} p{0.85\textwidth}}
\textbf{Arabic} & \textbf{Transliterations} \\ 
\hline
\textbf{\AR{أنا}} & ana~(20) \\ 
\textbf{\AR{إنسان}} & insan~(12), insane~(3), inssane~(3), in'ssen~(1) \\ 
\textbf{\AR{آتمنا}} & atamana~(2), atamanna~(2), atmana~(2), atmna~(2), netmenna~(2), netmna~(2), ntmena~(2), atemanna~(1), atmanna~(1), atmena~(1), atmenna~(1), etmena~(1), etmenna~(1) \\ 
\textbf{\AR{ماء}} & ma~(8), lma~(4), maa~(3), el ma~(1), elma~(1), l'ma~(1), ma'a~(1), ma2~(1), mae~(1) \\ 
\textbf{\AR{سؤال}} & sou2al~(7), so2al~(3), soual~(3), sou'al~(2), so2aol~(1), soaal~(1), soal~(1), souaal~(1) \\ 
\textbf{\AR{سلام}} & salam~(18), salaam~(1) \\ 
\textbf{\AR{برا}} & berra~(7), barra~(5), bera~(4), bara~(2), beraa~(1), bra~(1), brra~(1) \\ 
\textbf{\AR{ماكلة}} & makla~(17), maakla~(1) \\ 
\textbf{\AR{حمى}} & 7emma~(4), hemma~(4), 7ama~(3), 7amma~(2), 7oma~(2), 7ema~(1), 7ma~(1), hema~(1), l7ama~(1) \\ 
\textbf{\AR{بزاف}} & bezzaf~(8), bezaf~(4), bzaf~(4), bzf~(2), bezef~(1), bzeeeef~(1) \\ 
\textbf{\AR{بلاكار}} & placard~(12), placar~(3), blakar~(2), plakar~(2), blaker~(1) \\ 
\textbf{\AR{مثال}} & mithal~(11), mital~(3), methal~(2), mithale~(1), mithel~(1), mitèl~(1) \\ 
\textbf{\AR{جامع}} & jama3~(6), jame3~(5), djama3~(3), djame3~(3), jam3~(2), djamea~(1) \\ 
\textbf{\AR{حاجة}} & 7aja~(8), haja~(6), 7adja~(3), hadja~(3) \\ 
\textbf{\AR{صباح}} & sba7~(9), sbah~(5), sabah~(3), saba7~(2), ess'bah~(1), sebah~(1) \\ 
\textbf{\AR{خوك}} & khouk~(15), khok~(5), 5ouk~(1) \\ 
\textbf{\AR{هذاك}} & hadak~(11), hadhak~(3), 8adak~(2), hadek~(2), haadaak~(1), hadhek~(1) \\ 
\textbf{\AR{شجرة}} & chejra~(5), chajra~(4), chedjra~(3), chjra~(2), chajara~(1), chajarra~(1), chedjera~(1), sedj'ra~(1), shajra~(1), shejra~(1) \\ 
\textbf{\AR{رصاصة}} & rsasa~(11), rsassa~(3), ressassa~(2), r'ssassa~(1), rassassa~(1), rssasa~(1), rssassa~(1) \\ 
\textbf{\AR{ضحك}} & da7k~(7), dhak~(3), dahk~(2), dhahk~(2), d7ak~(1), d7ek~(1), dha7ak~(1), dha7k~(1), eddahk~(1), ldahk~(1) \\ 
\textbf{\AR{طبسي}} & tebsi~(5), tabsi~(4), 6ebsi~(2), tabssi~(2), tbsi~(2), 6ebssi~(1), tebsy~(1), tob'ssi~(1), tobsi~(1), tobssi~(1) \\ 
\textbf{\AR{ظلمة}} & dalma~(5), delma~(5), dhelma~(3), dholma~(2), dhalma~(1), dlma~(1), thalma~(1) \\ 
\textbf{\AR{غدوة}} & ghedwa~(12), ghadwa~(3), ghodwa~(2), ghdwa~(1), ghudwa~(1) \\ 
\textbf{\AR{فرحة}} & far7a~(6), fer7a~(6), ferha~(6), far'ha~(1), farha~(1), frha~(1) \\ 
\textbf{\AR{كرافاطة}} & cravata~(8), kravata~(7), caravata~(1), crafita~(1), crava6a~(1), cravate~(1), cravatta~(1), karafata~(1) \\ 
\textbf{\AR{رقيق}} & r9i9~(11), rgig~(2), r'qiq~(1), re9i9~(1), rekik~(1), reqiq~(1), rghigh~(1), rqiq~(1), rquiq~(1), rqyq~(1) \\ 
\textbf{\AR{قصبر}} & 9osbor~(5), 9osber~(2), kesbar~(2), kosbor~(2), qosbor~(2), 9esber~(1), 9esbor~(1), 9osbar~(1), 9sir~(1), 9ssber~(1), gousbr~(1), kassbar~(1), qosbar~(1) \\ 
\textbf{\AR{كوشة}} & koucha~(15), coucha~(2), kocha~(2), four~(1) \\ 
\textbf{\AR{هرب}} & hrab~(6), hreb~(5), 8reb~(3), ahrab~(1), h'rob~(1), harab~(1), harb~(1), herab~(1), herb~(1), herrab~(1) \\ 
\textbf{\AR{شابة‎}} & chaba~(9), chabba~(8), chebba~(2) \\ 
\textbf{\AR{جنينة الحومة}} & djeninat el homa~(1), djeninat el houma~(1), djinett el 7ouma~(1), djninat el 7ouma~(1), djninate alhouma~(1), djninet el7ouma~(1), djninet elhouma~(1), djninet l'houma~(1), jeninat el 7oma~(1), jnanat l7ouma~(1), jnina l7ouma~(1), jninat el 7ouma~(1), jninat el homa~(1), jninat l7ouma~(1), jninet alhouma~(1), jninet el 7ouma~(1), jninet l7ouma~(1), jniyna l7oma~(1) \\ 
\textbf{\AR{وحدي}} & wa7di~(7), wahdi~(7), we7di~(2), wehdi~(2), ouahdi~(1), wahdy~(1) \\ 
\textbf{\AR{عوام}} & 3awam~(9), 3awwam~(2), 3wam~(2), 3awan~(1), 3awem~(1), 3ewam~(1), 3ewwam~(1), 3oum~(1), 3ouwwam~(1), 3owam~(1), aouame~(1) \\ 
\textbf{\AR{روح}} & rouh~(7), rou7~(6), ro7~(5), roh~(2), reuh~(1) \\ 
\textbf{\AR{كولو}} & koulou~(11), koulo~(4), kolo~(2), kol~(1), kulu~(1) \\ 
\textbf{\AR{يقرا}} & ye9ra~(8), y9ra~(2), ya9ra~(2), yeqra~(2), yagra~(1), yakra~(1), yaqra~(1), yekra~(1), yekraa~(1) \\ 
\textbf{\AR{يستنى}} & yestenna~(7), yestena~(6), yestna~(2), ystna~(2), yasstana~(1), ystena~(1) \\ 
\textbf{\AR{يسقسي}} & yse9si~(3), ise9si~(2), ys9si~(2), ysa9si~(2), isseq'ssi~(1), isseqssi~(1), isseqssy~(1), yes9si~(1), yessakssi~(1), yesseksi~(1), ysa9ssi~(1), ysagsi~(1), yse9ssi~(1), yssa9si~(1) \\ 
\textbf{\AR{ياسمين}} & yasmine~(13), yasmin~(3), yaasmine~(1), yassmine~(1) \\ 
\textbf{\AR{جراچ}} & garage~(8), djeradj~(1), garaje~(1) \\
\textbf{\AR{تڜبتڜاق}} & tchebtcha9~(3), chbcha9~(2), chebcha9~(2), tchebcha9~(2), chebtchak~(1), tchabtcha9~(1), tchbcha9~(1), tchbtchag~(1), tcheb'tchaq~(1), tchebchak~(1), tchebchaq~(1), tchebtcha9a~(1), tchebtchaq~(1), techbtcha9~(1), this word is not a dish~(1) \\ 
\textbf{\AR{ڜكڜوكة}} & chekchouka~(6), chakchouka~(3), chkchouka~(2), tchekchouka~(2), tchektchouka~(2), chekchoka~(1), tchek'tchouka~(1), tchetchouka~(1) \\ 
\textbf{\AR{ڤليل}} & guellil~(7), glil~(3), guelil~(3), 9lil~(2), fa9ir~(1), flil~(1), ghelil~(1), ghellil~(1), guellile~(1) \\ 
\textbf{\AR{ڤايلة}} & gayla~(13), gaila~(1), gheyla~(1), guaila~(1), guayla~(1), gueyla~(1), l'ghayla~(1) \\ 
\textbf{\AR{زفيطي}} & zfiti~(7), zviti~(6), zefiti~(2), zvi6i~(2), z'ghiti~(1), zfity~(1), zvitti~(1) \\
\hline
\end{longtable}

\subsection{Egyptian words}
\begin{longtable}{p{0.08\textwidth} p{0.85\textwidth}}
\textbf{Arabic} & \textbf{Transliterations} \\ 
\hline
\textbf{\AR{أنا}} & ana~(22), 2na~(2) \\ 
\textbf{\AR{إنتماء}} & entema2~(10), entma2~(7), 2ntima2~(2), 2ntmaa2~(2), entemaa2~(1), entima2~(1), intima2~(1), intimaa2~(1) \\ 
\textbf{\AR{آمال}} & amaal~(9), amal~(8), 2amal~(5), 2amaal~(2), aamaal~(1) \\ 
\textbf{\AR{جزء}} & goz2~(20), juz'~(2), gooz2~(1), joz2~(1) \\ 
\textbf{\AR{لئيم}} & la2eem~(17), l2eem~(2), la'eem~(2), l2em~(1), l2im~(1), la2em~(1), la2ym~(1) \\ 
\textbf{\AR{سؤال}} & so2al~(25) \\ 
\textbf{\AR{استعمال}} & este3mal~(11), est3mal~(8), 2st3mal~(2), isti3mal~(2), esti3maal~(1), iste3maal~(1) \\ 
\textbf{\AR{استعملوا}} & esta3melo~(10), est3mlo~(8), 2st3mlo~(2), 2esta3milo~(1), est3melo~(1), esta3malo~(1), esta3meloh~(1), ista3malo~(1) \\ 
\textbf{\AR{سلام}} & salam~(25) \\ 
\textbf{\AR{حمى}} & 7omma~(14), 7oma~(6), 7ama~(1), 7ommah~(1), 7ooma~(1), homa~(1), homma~(1) \\ 
\textbf{\AR{باب}} & bab~(20), baab~(5) \\ 
\textbf{\AR{باركيه}} & barkeh~(7), barkeeh~(4), parkeeh~(3), parkeh~(3), parquet~(3), barkeih~(1), barkih~(1), barkyh~(1), barqueh~(1), parke~(1) \\ 
\textbf{\AR{ثانوية}} & sanawya~(5), thanaweya~(5), thanawya~(3), thanwya~(3), sanaweya~(2), sanaweyah~(2), thanaweyya~(2), sanwya~(1), thanawia~(1), thanawy~(1) \\ 
\textbf{\AR{ثعبان}} & tho3ban~(9), te3ban~(8), so3ban~(7), te3baan~(1) \\ 
\textbf{\AR{جمال}} & gamal~(21), gamaal~(2), jamal~(2) \\ 
\textbf{\AR{جراج}} & garage~(19), garag~(4), garaj~(1), garash~(1) \\ 
\textbf{\AR{حاجة}} & 7aga~(18), haga~(6), 7ga~(1) \\ 
\textbf{\AR{صباح}} & saba7~(16), sabah~(7), saha7~(1), sba7~(1) \\ 
\textbf{\AR{خمسة}} & 5amsa~(17), khamsa~(7), khamsah~(1) \\ 
\textbf{\AR{ذئب}} & ze2b~(15), deeb~(6), th'eeb~(1), the2b~(1), z2b~(1), zi2b~(1) \\ 
\textbf{\AR{شمس}} & shams~(23), 4ams~(2) \\ 
\textbf{\AR{صاحبي}} & sa7bi~(11), sa7by~(8), sahbi~(3), sahby~(3) \\ 
\textbf{\AR{ضحك}} & de7k~(18), d7k~(4), dehk~(2), dahak~(1) \\ 
\textbf{\AR{طيران}} & tayaran~(20), tayran~(3), 6ayaran~(1), tyaran~(1) \\ 
\textbf{\AR{ظلم}} & zolm~(24), tholm~(1) \\ 
\textbf{\AR{غياب}} & gheyab~(9), 3'yab~(3), 8eyab~(3), ghiyab~(3), 8yab~(2), '3eyaab~(1), 3'eyab~(1), 3'iab~(1), gheab~(1), ghyab~(1) \\ 
\textbf{\AR{فراولة}} & farawla~(18), farwla~(4), frawla~(2), farawlla~(1) \\ 
\textbf{\AR{فارندا}} & varanda~(19), faranda~(2), farnda~(1), phranda~(1), varnda~(1), veranda~(1) \\ 
\textbf{\AR{قانون}} & qanoon~(13), qanon~(5), kanoon~(3), kanon~(2), qanun~(1), qanuun~(1) \\ 
\textbf{\AR{بتقول}} & bet2ol~(8), bet2ool~(7), bt2ol~(4), bt2ool~(2), bet'ool~(1), bet2ul~(1), betqool~(1), bt2oul~(1) \\ 
\textbf{\AR{كتاب}} & ketab~(18), kitab~(5), ktaab~(1), ktab~(1) \\ 
\textbf{\AR{هايل}} & hayel~(20), hayl~(3), haiel~(1), hayeel~(1) \\ 
\textbf{\AR{جنينة}} & genena~(9), geneena~(4), genenah~(3), gnena~(3), gneena~(2), geneynah~(1), genina~(1), gnenaa~(1), gnina~(1) \\ 
\textbf{\AR{جنينة الورد}} & genenet el ward~(3), genena~(2), genent elward~(2), geneenat alward~(1), geneenet el ward~(1), geneenet elward~(1), genena el ward~(1), genenat alward~(1), genenet elward~(1), genent el ward~(1), genent elwrd~(1), geninet elward~(1), gneenat el ward~(1), gneenet el ward~(1), gnenat el ward~(1), gnenat ward~(1), gnenet el ward~(1), gnenet l ward~(1), gnenet ward~(1), gnent el ward~(1), gnint ward~(1) \\ 
\textbf{\AR{ألوان}} & alwan~(19), 2lwan~(3), 2alwan~(1), alwaan~(1) \\ 
\textbf{\AR{معروف}} & ma3roof~(14), ma3rof~(3), m3roof~(2), ma3rouf~(2), m3arof~(1), m3rof~(1), ma'rof~(1), ma3ruf~(1) \\ 
\textbf{\AR{نمو}} & nemow~(4), nemw~(3), nomow~(3), nmow~(2), nomo~(2), nomu~(2), nemew~(1), nemmo~(1), nemoww~(1), nmoo~(1), nomou~(1), nomoww~(1), nomw~(1), numow~(1), numuw~(1) \\ 
\textbf{\AR{وادي}} & wadi~(13), wady~(9), waadi~(1), waady~(1), wadii~(1) \\ 
\textbf{\AR{أبو ...}} & abo~(17), abu~(6), 2boo~(1), abou~(1) \\ 
\textbf{\AR{ياسمين}} & yasmin~(14), yasmeen~(6), yasmine~(5) \\ 
\textbf{\AR{جراچ‎}} & garage~(20), garag~(4), garash~(1) \\ 
\textbf{\AR{ڤارندا}} & varanda~(21), veranda~(2), varnda~(1) \\ 
\hline
\end{longtable}

\subsection{Lebanese words}
\begin{longtable}{p{0.08\textwidth} p{0.85\textwidth}}
\textbf{Arabic} & \textbf{Transliterations} \\ 
\hline
\textbf{\AR{أمل}} & amal~(12), 2amal~(2) \\ 
\textbf{\AR{إسم}} & esem~(9), 2essem~(2), esm~(2), isim~(1) \\ 
\textbf{\AR{إيام}} & iyyem~(6), iyem~(2), 2eyem~(1), 2iyem~(1), ayam~(1), eyyem~(1), iyam~(1), iyyeim~(1) \\ 
\textbf{\AR{آخر}} & ekher~(3), ekhir~(3), e5ir~(2), 2a5ar~(1), 2e5er~(1), 2ei5er~(1), 2ekher~(1), akhar~(1), e5er~(1) \\ 
\textbf{\AR{جزء}} & jeze2~(13), gizi'~(1) \\ 
\textbf{\AR{رئيس}} & ra2is~(11), ra2iss~(2), raeiss~(1) \\ 
\textbf{\AR{سؤال}} & sou2el~(5), so2al~(3), sou2al~(3), su2al~(2), sou'al~(1) \\ 
\textbf{\AR{استعمال}} & este3mel~(8), 2este3mel~(2), esteemel~(1), iste3mal~(1), iste3mel~(1), isti3mal~(1) \\ 
\textbf{\AR{سلام}} & salam~(9), salem~(5) \\ 
\textbf{\AR{استعملوا}} & sta3malo~(3), sta3malou~(2), staamalo~(2), 2esta3malou~(1), 2esta3mlou~(1), esta3emlo~(1), esta3malou~(1), estaamalo~(1), ista3melo~(1), sta3emlo~(1) \\ 
\textbf{\AR{فتوى}} & fatwa~(13), fatwe~(1) \\ 
\textbf{\AR{باب}} & beb~(12), bab~(1), beib~(1) \\ 
\textbf{\AR{بيتزا}} & pizza~(14) \\ 
\textbf{\AR{ثواني}} & sawene~(8), sawani~(2), saweni~(2), thawani~(2) \\ 
\textbf{\AR{ثورة}} & sawra~(8), thawra~(5), sarwa~(1) \\ 
\textbf{\AR{دجاج}} & djej~(8), dajaj~(2), jej~(2), djaj~(1), djeij~(1) \\ 
\textbf{\AR{جراج}} & garage~(5), jaraj~(1), jraj~(1) \\ 
\textbf{\AR{حاج}} & 7aj~(7), haj~(4), 7aje~(1), 7ej~(1), haje~(1) \\ 
\textbf{\AR{صباح}} & saba7~(7), sabah~(6), sebe7~(1) \\ 
\textbf{\AR{خمسة}} & khamse~(6), 5amse~(5), 5amsi~(1), khamsa~(1), khamseh~(1) \\ 
\textbf{\AR{ذكي}} & zake~(7), zaki~(7) \\ 
\textbf{\AR{شاورما}} & shawarma~(11), chawarma~(1), chawerma~(1), shawerma~(1) \\ 
\textbf{\AR{صحتين}} & sa7ten~(6), sahtein~(5), sahten~(2), sa7tein~(1), sahtayn~(1) \\ 
\textbf{\AR{مضطر}} & medtar~(8), mettar~(3), modtar~(2), mitdar~(1) \\ 
\textbf{\AR{طاولة}} & tawla~(6), tawle~(5), tawleh~(2), tewla~(1) \\ 
\textbf{\AR{مظاهرة}} & mouzahara~(10), mozahara~(2), muzahara~(2) \\ 
\textbf{\AR{غريب}} & gharib~(9), 8arib~(5), ghareeb~(1) \\ 
\textbf{\AR{فلافل}} & falafel~(7), falefel~(4), falefil~(3) \\ 
\textbf{\AR{رقيق}} & r2i2~(8), rakik~(3), r2e2~(1), ra2i2~(1), rkik~(1) \\ 
\textbf{\AR{قديم}} & adim~(9), 2adim~(4), kadeem~(1) \\ 
\textbf{\AR{بكير}} & bakkir~(7), bakir~(6), bakeer~(1) \\ 
\textbf{\AR{هوا}} & hawa~(14) \\ 
\textbf{\AR{شمعة}} & cham3a~(5), sham3a~(5), shamaa~(3), chamaa~(1) \\ 
\textbf{\AR{حديقة الورد}} & 7ade2et wared~(1), 7adi2a~(1), 7adi2et el ward~(1), 7adi2et el wared~(1), 7adi2et wared~(1), 7adi2it el wared~(1), hadi2a~(1), hadi2et l ward~(1), hadi2et wroud~(1), hadi2it el ward~(1), hadi2it l wared~(1), hadika~(1), hadiqat elward~(1), jnaynet l wared~(1) \\ 
\textbf{\AR{ألوان}} & alwen~(8), alwan~(3), 2alwen~(2), alwein~(1) \\ 
\textbf{\AR{وحدي}} & wa7di~(5), wa7de~(4), wahde~(4), lawahde~(1) \\ 
\textbf{\AR{معروف}} & ma3rouf~(10), maarouf~(4) \\ 
\textbf{\AR{جدو}} & jeddo~(11), jedo~(2), jedde~(1) \\ 
\textbf{\AR{حلو}} & helo~(6), 7elo~(4), 7elou~(3), hilu~(1) \\ 
\textbf{\AR{ياسمين}} & yasmine~(7), yasmin~(6), yesmin~(1) \\ 
\textbf{\AR{ڤيزا}} & visa~(13), vizza~(1) \\ 
\hline
\end{longtable}

\subsection{Moroccan words}
\begin{longtable}{p{0.08\textwidth} p{0.85\textwidth}}
\textbf{Arabic} & \textbf{Transliterations} \\ 
\hline
\textbf{\AR{أسئلة}} & as2ila~(15), 2ass2ila~(1), ass2ila~(1), assela~(1), assila~(1) \\ 
\textbf{\AR{إنسان}} & insan~(11), inssan~(6), insane~(3), 2nsan~(1), inssane~(1) \\ 
\textbf{\AR{آخر}} & akhir~(10), akher~(4), akhor~(2), 2akhir~(1), akhr~(1) \\ 
\textbf{\AR{جزء}} & jouz2~(10), joz2~(4), jouzeae~(1), jozae~(1) \\ 
\textbf{\AR{رئيس}} & ra2iss~(12), ra2is~(5), raiss~(4), ra-iss~(1) \\ 
\textbf{\AR{سؤال}} & sou2al~(11), so2al~(5), soual~(2), soael~(1), sou-al~(1), su2al~(1) \\ 
\textbf{\AR{استعمال}} & isti3mal~(11), sti3mal~(7), istiaeamal~(1) \\ 
\textbf{\AR{استعملوا}} & ste3mlou~(5), ista3malou~(3), st3mlo~(2), sta3mlo~(2), asta3mloua~(1), ista3milo~(1), ista3mlo~(1), istaeamalou~(1), st3mlou~(1), sta3mlou~(1), steemlu~(1) \\ 
\textbf{\AR{تلفازا}} & tlfaza~(8), telfaza~(4), talfaza~(1), tlfasa~(1), tlvaza~(1) \\ 
\textbf{\AR{فتوى}} & fatwa~(18), ftwa~(2), fetwa~(1) \\ 
\textbf{\AR{باب}} & bab~(22) \\ 
\textbf{\AR{بلاكار}} & placard~(14), placar~(3), plakar~(2), plaqar~(1) \\ 
\textbf{\AR{مثال}} & mital~(14), mithal~(2), mitale~(1) \\ 
\textbf{\AR{شرجم}} & cherjem~(12), charjam~(4), chrjm~(3), charjem~(1), chrjem~(1) \\ 
\textbf{\AR{حمام}} & 7emmam~(4), hamam~(4), hammam~(4), 7ammam~(3), 7mam~(2), hemmam~(2), 7amam~(1), 7emam~(1) \\ 
\textbf{\AR{صباح}} & sbah~(10), sba7~(9), saba7~(3), sabah~(1), sbah7~(1) \\ 
\textbf{\AR{سخون}} & skhoun~(15), skhoune~(3), skhun~(2), skhn~(1), skhon~(1) \\ 
\textbf{\AR{مذاق}} & mada9~(16), madaq~(5), madaque~(1) \\ 
\textbf{\AR{مشكيل}} & mochkil~(8), mouchkil~(8), mouchkila~(4), muchkil~(1) \\ 
\textbf{\AR{خلص}} & kheless~(4), khalass~(3), khelles~(3), khelless~(3), khallass~(1), khalles~(1), khalless~(1), khalss~(1), khellass~(1), khlas~(1), khles~(1), khless~(1), khlss~(1) \\ 
\textbf{\AR{عضلة}} & 3adala~(15), 3edla~(1), aeadala~(1) \\ 
\textbf{\AR{طيارة}} & tiyara~(6), tyara~(5), teyara~(3), tayara~(2), tayyara~(2), tiara~(2), tyyara~(2), teyyara~(1) \\ 
\textbf{\AR{ظلم}} & dolm~(12), doulm~(4), dalm~(1), ddolm~(1), delm~(1), dhlm~(1), dholm~(1) \\ 
\textbf{\AR{غيام}} & ghyam~(17), ghiam~(1), ghiyam~(1), ghyoum~(1) \\ 
\textbf{\AR{فران}} & ferran~(9), fran~(5), farrane~(2), feran~(2), ferrane~(2), faran~(1), frran~(1) \\ 
\textbf{\AR{كرافاطا}} & cravata~(12), kravata~(3), cravate~(2), grafata~(2), crafata~(1), gravata~(1) \\ 
\textbf{\AR{رقيق}} & r9i9~(14), rqiq~(5), raquique~(1), re9i9~(1) \\ 
\textbf{\AR{قال}} & 9al~(9), gal~(8), qal~(1), qala~(1), qual~(1) \\ 
\textbf{\AR{بكري}} & bekri~(12), bkri~(6), bakri~(3), bqri~(1) \\ 
\textbf{\AR{كركاع}} & garga3~(6), grga3~(5), guerga3~(5), gerga3~(4), gargaea~(1), ge3ga3~(1), gurgua3~(1) \\ 
\textbf{\AR{هرب}} & hreb~(15), hrab~(5), hrb~(2) \\ 
\textbf{\AR{هدية}} & hadiya~(5), hadia~(3), hdiya~(2), 8adiya~(1), hadeyyat~(1), hadiyya~(1), hdia~(1) \\ 
\textbf{\AR{هدية العيد}} & hadiyat l3id~(2), 8adiyat al eid~(1), hadeyyat el3id~(1), hadiat laid~(1), hadiyat l3eid~(1), hadiyat l’3id~(1), hdiat l3id~(1), hdiyt l3id~(1) \\ 
\textbf{\AR{ألوان}} & alwan~(18), alouane~(1), alwane~(1) \\ 
\textbf{\AR{روز}} & roz~(12), rouz~(6), rouze~(1), rroz~(1) \\ 
\textbf{\AR{وردة}} & warda~(10), werda~(8), wrda~(5) \\ 
\textbf{\AR{يشربو}} & icherbou~(3), ycherbo~(3), ychrbo~(3), ycherbou~(2), yachrbo~(1), yachribou~(1), ycharbo~(1), ychrbou~(1), ychrbu~(1), yoshorbou~(1) \\ 
\textbf{\AR{بياع}} & biya3~(8), bya3~(6), baya3~(2), beyaea~(1), biyya3~(1) \\ 
\textbf{\AR{خالي}} & khali~(18) \\ 
\textbf{\AR{ياسمين}} & yasmine~(14), yasmin~(4), yassmin~(2), yassmine~(1) \\ 
\textbf{\AR{ڤيلا}} & villa~(21) \\ 
\textbf{\AR{ڭيطار}} & guitar~(10), guitare~(8), gitar~(2), guitara~(1) \\ 
\hline
\end{longtable}

\subsection{Tunisian words}
\begin{longtable}{p{0.08\textwidth} p{0.85\textwidth}}
\textbf{Arabic} & \textbf{Transliterations} \\ 
\hline
\textbf{\AR{أنا}} & ena~(13), ana~(8), eni~(6), ane~(1), eny~(1) \\ 
\textbf{\AR{إستنا}} & estana~(12), estanna~(6), stana~(4), stanna~(3), istana~(2), esstena~(1), estena~(1), istanna~(1) \\ 
\textbf{\AR{آخر}} & e5er~(7), ekher~(7), akher~(6), a5er~(3), a5ar~(2), a5ra~(1), ache~(1), akhar~(1), akhér~(1), lakher~(1) \\ 
\textbf{\AR{ماء}} & mee~(7), me~(6), ma~(5), ma2~(3), maa~(2), me2~(2), mé~(2), maa2~(1), met~(1), mi~(1) \\ 
\textbf{\AR{رئيس}} & ra2is~(21), ra2iss~(3), rais~(2), r2is~(1), raeiss~(1), raiis~(1), rayis~(1) \\ 
\textbf{\AR{سؤال}} & sou2el~(17), souel~(5), sou2al~(3), sou2l~(2), so2el~(1), soual~(1) \\ 
\textbf{\AR{استعمال}} & este3mel~(9), isti3mel~(4), esta3mel~(2), este3mal~(2), esti3mel~(2), ista3mal~(2), ista3mel~(2), iste3mel~(2), est3mel~(1), esta3mal~(1), estaamel~(1), esti3mal~(1), istaamel~(1), istamel~(1) \\ 
\textbf{\AR{يلعبوا}} & yal3bou~(16), yala3bou~(6), yal3abou~(2), yal3bo~(1), yala3bo~(1), yalaabo~(1), yalaabou~(1), yale3bou~(1), yaleebou~(1) \\ 
\textbf{\AR{طيار}} & tayar~(19), tayyar~(6), pilot~(1), tayara~(1) \\ 
\textbf{\AR{حمى}} & 7omma~(9), 7oma~(8), hamma~(2), homma~(2), 7amya~(1), 7ouma~(1), hama~(1) \\ 
\textbf{\AR{باب}} & beb~(28), bab~(1), baba~(1) \\ 
\textbf{\AR{بلاكار}} & placard~(12), blakar~(5), plakar~(5), placar~(4), bla9ar~(1), blacare~(1), blakra~(1), placcard~(1), plakard~(1) \\ 
\textbf{\AR{ثما}} & thama~(15), thamma~(6), famma~(3), fama~(2), thema~(2), them~(1), thma~(1) \\ 
\textbf{\AR{جامع}} & jama3~(10), jame3~(9), jema3~(9), jamaa~(1), jemaa~(1), jeme3~(1) \\ 
\textbf{\AR{حاجة}} & 7aja~(19), haja~(10), heja~(1) \\ 
\textbf{\AR{صباح}} & sbe7~(10), sba7~(6), sbeh~(6), sbah~(4), saba7~(1), sbeeh~(1) \\ 
\textbf{\AR{خوك}} & 5ouk~(16), khouk~(13), 5hou~(1) \\ 
\textbf{\AR{هذاكا}} & hadhaka~(10), hedheka~(8), hadheka~(4), hatheka~(3), hedhaka~(2), ha4aka~(1), hetheka~(1) \\ 
\textbf{\AR{برشا}} & barcha~(23), barsha~(5), brcha~(1) \\ 
\textbf{\AR{صاحبي}} & sa7bi~(19), sahbi~(7), sa7by~(1), sahby~(1), sahib~(1) \\ 
\textbf{\AR{ضحك}} & dha7k~(6), dha7ek~(3), dho7k~(3), dh7ak~(2), dhohk~(2), 4a7ek~(1), dahek~(1), dha7ka~(1), dhahek~(1), dhahik~(1), dhaka~(1), dhe7ek~(1), dhe7k~(1), dhhak~(1), dhhek~(1), dhi7k~(1), dho7ek~(1), dhohek~(1), yadhhak~(1), yadhhek~(1) \\ 
\textbf{\AR{يعطي}} & ya3ti~(22), yaati~(5), ya3ty~(2) \\ 
\textbf{\AR{يظهرلي}} & yodhhorli~(10), yodhherli~(3), yodhorli~(3), dhaherli~(2), yodh'horli~(2), yedhorli~(1), yethhorli~(1), yo4horli~(1), yodh horli~(1), yodhehorli~(1), yodhherly~(1), yodhhrli~(1), yothhourli~(1), youdhhorli~(1), youdhohorli~(1) \\ 
\textbf{\AR{غدوة}} & ghodwa~(18), 8odwa~(8), ghodoi~(2), 4odwa~(1), 8odoi~(1), ghoudwa~(1) \\ 
\textbf{\AR{فلوس}} & flous~(24), flouss~(5), flows~(1) \\ 
\textbf{\AR{فيراندا}} & viranda~(27), firanda~(2), veranda~(1), véranda~(1) \\ 
\textbf{\AR{قديم}} & 9dim~(18), gdim~(7), kdim~(2), 9adim~(1), gdime~(1) \\ 
\textbf{\AR{كتاب}} & kteb~(22), kiteb~(3), ktab~(2), kteeb~(1), ktéb~(1) \\ 
\textbf{\AR{هكا}} & haka~(18), hakka~(7), heka~(4), hika~(1) \\ 
\textbf{\AR{باهية}} & behya~(14), behia~(4), bahia~(2), bahya~(2), behiya~(2), bahi~(1), bahiya~(1), behi~(1), behyaa~(1), behyia~(1), bhy~(1) \\ 
\textbf{\AR{جنينة الحومة}} & jninet el 7ouma~(9), jninet l 7ouma~(3), jninet el houma~(2), jninet lhouma~(2), jninat el7ouma~(1), jninet 7ouma~(1), jninet el homa~(1), jninet el7ouma~(1), jninet houma~(1), jninet il7oma~(1), jninit alhouma~(1), jnint l7ouma~(1), jninét elhouma~(1), joint l7ouma~(1) \\ 
\textbf{\AR{ألوان}} & alwen~(24), alwan~(4), alwén~(1) \\ 
\textbf{\AR{معروف}} & ma3rouf~(24), maarouf~(5), marouf~(1) \\ 
\textbf{\AR{وحدي}} & wa7di~(19), wahdi~(8), wahdy~(1) \\ 
\textbf{\AR{بو}} & bou~(29) \\ 
\textbf{\AR{ولدي}} & weldi~(27), waldy~(1), wildi~(1) \\ 
\textbf{\AR{ياسمين}} & yasmine~(16), yasmin~(9), yassmine~(4), yesmine~(1) \\ 
\textbf{\AR{يمشي}} & yemchi~(22), yemshi~(4), yamchi~(2), yemchy~(1), yimchi~(1) \\ 
\textbf{\AR{يڤدم}} & yegdem~(17), ygadem~(6), egaddem~(1), y9adem~(1), yegdm~(1), yeghdem~(1), ygadam~(1), ygaddem~(1), ygadém~(1) \\ 
\textbf{\AR{ڥيلا}} & villa~(28), vela~(1), vila~(1) \\ 
\hline
\end{longtable}

\end{document}

%% file: exp2_results.tex
\setlength{\tabcolsep}{2.5pt}
\resizebox{\textwidth}{!}{%
\begin{tabular}{lcc|ccc|ccc|ccc|ccc|ccc}
\multicolumn{2}{c}{\multirow{3}{*}{\textbf{Split}}} &  \multirow{3}{*}{\textit{\textbf{N}}} & & \multicolumn{2}{c}{\textbf{Algeria}} & & \multicolumn{2}{c}{\textbf{Egypt}} & & \multicolumn{2}{c}{\textbf{Lebanon}} & & \multicolumn{2}{c}{\textbf{Morocco}} & & \multicolumn{2}{c}{\textbf{Tunisia}}\\
& & & \textbf{\textit{N\textsubscript{Com}}} & {\symbolafont \symbol{"1F464}}\textsubscript{\textbf{A}} & {\symbolafont \symbol{"1F464}}\textsubscript{\textbf{B}} & \textbf{\textit{N\textsubscript{Com}}} & {\symbolafont \symbol{"1F464}}\textsubscript{\textbf{C}} & {\symbolafont \symbol{"1F464}}\textsubscript{\textbf{D}} & \textbf{\textit{N\textsubscript{Com}}} & {\symbolafont \symbol{"1F464}}\textsubscript{\textbf{E}} & {\symbolafont \symbol{"1F464}}\textsubscript{\textbf{F}} & \textbf{\textit{N\textsubscript{Com}}} & {\symbolafont \symbol{"1F464}}\textsubscript{\textbf{G}} & {\symbolafont \symbol{"1F464}}\textsubscript{\textbf{H}} & \textbf{\textit{N\textsubscript{Com}}} & {\symbolafont \symbol{"1F464}}\textsubscript{\textbf{I}} & {\symbolafont \symbol{"1F464}}\textsubscript{\textbf{J}}\\
& & & & \checkmark/?/\ding{55} & \checkmark/?/\ding{55} & & \checkmark/?/\ding{55} & \checkmark/?/\ding{55} & & \checkmark/?/\ding{55} & \checkmark/?/\ding{55} & & \checkmark/?/\ding{55} & \checkmark/?/\ding{55} & & \checkmark/?/\ding{55} & \checkmark/?/\ding{55} \\
\midrule

\multirow{2}{*}{\textbf{Algeria}} & {\symbolafont \symbol{"1F4C3}}\textsubscript{A} & \multirow{2}{*}{112} & \multirow{2}{*}{112} & {\sameCountryCorrect{{105}/3/4}} & {\sameCountryIncorrect{{87}/0/25}} & \multirow{2}{*}{34} & {\diffCountryDiff{3/2/29}} & \diffCountrySomeSim{5/22/7} & \multirow{2}{*}{50} & {\diffCountryDiff{2/9/39}} & {\diffCountryDiff{4/3/43}} & \multirow{2}{*}{46} & {\diffCountryDiff{9/1/36}} & \diffCountrySim{38/8/0} & \multirow{2}{*}{47} & {\diffCountryDiff{13/1/33}} & \diffCountrySomeSim{18/16/13} \\
 & {\symbolafont \symbol{"1F4C3}}\textsubscript{B} &  &  & {\sameCountryIncorrect{{100}/3/9}} & {\sameCountryCorrect{{108}/1/3}} &  & {\diffCountryDiff{3/0/31}} & {\diffCountryDiff{3/9/22}} &  & {\diffCountryDiff{8/7/35}} & {\diffCountryDiff{6/5/39}} &  & {\diffCountryDiff{7/5/34}} & \diffCountrySim{38/8/0} &  & {\diffCountryDiff{8/5/34}} & \diffCountrySomeSim{17/16/14} \\
 &  &  &  &  &  &  &  &  &  &  &  &  &  &  &  &  \\
\multirow{2}{*}{\textbf{Egypt}} & {\symbolafont \symbol{"1F4C3}}\textsubscript{C} & \multirow{2}{*}{104} & \multirow{2}{*}{34} & {\diffCountryDiff{7/0/27}} & {\diffCountryDiff{7/0/27}} & \multirow{2}{*}{104} & {\sameCountryCorrect{{104}/0/0}} & {\sameCountryCorrect{{102}/0/2}} & \multirow{2}{*}{87} & \diffCountrySomeSim{38/11/38} & \diffCountrySomeSim{35/10/42} & \multirow{2}{*}{12} & {\diffCountryDiff{1/0/11}} & \diffCountrySomeSim{6/1/5} & \multirow{2}{*}{19} & {\diffCountryDiff{4/0/15}} & {\diffCountryDiff{3/2/14}} \\
 & {\symbolafont \symbol{"1F4C3}}\textsubscript{D} &  &  & {\diffCountryDiff{8/1/25}} & {\diffCountryDiff{10/0/24}} &  & {\sameCountryCorrect{{99}/1/4}} & {\sameCountryCorrect{{102}/1/1}} &  & \diffCountrySomeSim{35/9/43} & {\diffCountryDiff{26/12/49}} &  & {\diffCountryDiff{2/0/10}} & \diffCountrySomeSim{6/1/5} &  & {\diffCountryDiff{8/0/11}} & \diffCountrySomeSim{4/7/8} \\
 &  &  &  &  &  &  &  &  &  &  &  &  &  &  &  &  \\
\multirow{2}{*}{\textbf{Lebanon}} & {\symbolafont \symbol{"1F4C3}}\textsubscript{E} & \multirow{2}{*}{122} & \multirow{2}{*}{50} & {\diffCountryDiff{7/1/42}} & {\diffCountryDiff{8/1/41}} & \multirow{2}{*}{87} & {\diffCountryDiff{17/5/65}} & {\diffCountryDiff{31/6/50}} & \multirow{2}{*}{122} & {\sameCountryCorrect{{121}/1/0}} & {\sameCountryCorrect{{110}/9/3}} & \multirow{2}{*}{16} & {\diffCountryDiff{3/3/10}} & \diffCountrySomeSim{7/3/6} & \multirow{2}{*}{31} & {\diffCountryDiff{5/0/26}} & {\diffCountryDiff{6/6/19}} \\
 & {\symbolafont \symbol{"1F4C3}}\textsubscript{F} &  &  & {\diffCountryDiff{5/0/45}} & {\diffCountryDiff{5/1/44}} &  & {\diffCountryDiff{6/2/79}} & {\diffCountryDiff{16/1/70}} &  & {\sameCountryCorrect{{121}/1/0}} & {\sameCountryCorrect{{122}/0/0}} &  & {\diffCountryDiff{1/0/15}} & \diffCountrySomeSim{4/5/7} &  & {\diffCountryDiff{3/0/28}} & {\diffCountryDiff{2/7/22}} \\
 &  &  &  &  &  &  &  &  &  &  &  &  &  &  &  &  \\
\multirow{2}{*}{\textbf{Morocco}} & {\symbolafont \symbol{"1F4C3}}\textsubscript{G} & \multirow{2}{*}{55} & \multirow{2}{*}{46} & \diffCountrySim{37/2/7} & \diffCountrySim{41/0/5} & \multirow{2}{*}{12} & {\diffCountryDiff{2/1/9}} & \diffCountrySomeSim{3/5/4} & \multirow{2}{*}{16} & {\diffCountryDiff{1/2/13}} & {\diffCountryDiff{1/0/15}} & \multirow{2}{*}{55} & {\sameCountryCorrect{{54}/1/0}} & {\sameCountryCorrect{{52}/3/0}} & \multirow{2}{*}{21} & {\diffCountryDiff{4/5/12}} & \diffCountrySim{14/3/4} \\
 & {\symbolafont \symbol{"1F4C3}}\textsubscript{H} &  &  & \diffCountrySim{38/1/7} & \diffCountrySim{37/1/8} &  & {\diffCountryDiff{2/0/10}} & \diffCountrySomeSim{4/4/4} &  & {\diffCountryDiff{4/2/10}} & {\diffCountryDiff{3/1/12}} &  & {\sameCountryCorrect{{50}/5/0}} & {\sameCountryCorrect{{52}/3/0}} &  & \diffCountrySomeSim{10/1/10} & \diffCountrySim{15/5/1} \\
 &  &  &  &  &  &  &  &  &  &  &  &  &  &  &  &  \\
\multirow{2}{*}{\textbf{Tunisia}} & {\symbolafont \symbol{"1F4C3}}\textsubscript{I} & \multirow{2}{*}{60} & \multirow{2}{*}{47} & \diffCountrySim{29/4/14} & \diffCountrySomeSim{23/3/21} & \multirow{2}{*}{19} & {\diffCountryDiff{6/1/12}} & \diffCountrySomeSim{5/7/7} & \multirow{2}{*}{31} & {\diffCountryDiff{8/1/22}} & {\diffCountryDiff{5/2/24}} & \multirow{2}{*}{21} & \diffCountrySomeSim{6/5/10} & \diffCountrySim{15/3/3} & \multirow{2}{*}{60} & {\sameCountryCorrect{{56}/0/4}} & {\sameCountryIncorrect{{47}/7/6}} \\
 & {\symbolafont \symbol{"1F4C3}}\textsubscript{J} &  &  & {\diffCountryDiff{15/4/28}} & {\diffCountryDiff{14/2/31}} &  & {\diffCountryDiff{2/0/17}} & \diffCountrySomeSim{4/6/9} &  & {\diffCountryDiff{9/0/22}} & {\diffCountryDiff{5/0/26}} &  & {\diffCountryDiff{4/3/14}} & \diffCountrySomeSim{8/10/3} &  & {\sameCountryIncorrect{{52}/4/4}} & {\sameCountryCorrect{{58}/2/0}} \\

\bottomrule
\end{tabular}%
}

%% file: algorithm.tex
\begin{figure*}[!phtb]
\centering
\begin{lstlisting}[caption={Function used to align Arabic words with their transliterations.},
                   label={lst:align_word},
                   basicstyle=\ttfamily\footnotesize,
                   numbers=left,
                   frame=single,
                   breaklines=true]
def align_word(arabic_word, transliterated_word):
    """
    Determine the longest alignment between an Arabic word and its transliterated form.
    This function uses a recursive approach to match letters based on a provided mapping.
    The letters in SKIPPABLE_LETTERS are vowels that can be present in Arabizi without a corresponding Arabic letter.
    """

    if not arabic_word:
        if not transliterated_word:
            # Both words are empty, return a match
            return 0, []

        # Skip vowels at the end (if needed)
        if transliterated_word[0] in SKIPPABLE_LETTERS:
            len_matched, remaining_alignment = align_word(
                arabic_word, transliterated_word[1:]
            )

            # If we can skip the vowel and still have a match, return it
            if len_matched >= 0:
                return len_matched, [("", transliterated_word[0])] + remaining_alignment

            # If we reach here, it means we have an Arabic word but no transliterated match
            return -inf, []

        # If we have an empty Arabic word but a transliterated word with a letter that cannot be skipped
        return -inf, []

    current_arabic_letter, remaining_word = arabic_word[0], arabic_word[1:]
    potential_letter_mappings = letters_map.get(current_arabic_letter, [])
    max_len, max_alignment = -inf, []

    # Try to match the current Arabic letter with all its potential transliterated forms based on the letters_map
    for matching in potential_letter_mappings:
        if transliterated_word.startswith(matching):
            # Attempt this matching
            len_matched, remaining_alignment = align_word(
                remaining_word, transliterated_word[len(matching) :]
            )

            # If we have a valid match, check if it is the longest found so far
            if len_matched >= 0:
                if len(matching) + len_matched > max_len:
                    max_len = len(matching) + len_matched
                    max_alignment = [
                        (current_arabic_letter, matching)
                    ] + remaining_alignment

    # Try skipping vowels without consuming from the Arabic word!
    if transliterated_word and transliterated_word[0] in SKIPPABLE_LETTERS:
        len_matched, remaining_alignment = align_word(
            arabic_word, transliterated_word[1:]
        )
        if len_matched >= 0:
            if len_matched > max_len:
                max_len = len_matched
                max_alignment = [("", transliterated_word[0])] + remaining_alignment

    return max_len, max_alignment
\end{lstlisting}
\end{figure*}

%% file: letter_alignment.tex
\begin{table*}[tbph]
\scriptsize
\caption{Aligned Arabic/Latin letters, aggregated across different words and participants. Each cell indicate the Latin transcription of an Arabic letter, with the transcription's frequency shown (between parentheses). \textbf{Note:} Mappings not included in the Wikipedia table are \textbf{bolded}, $\epsilon$ is an empty string, and sp. is a space. \taha{Generate the Table based on the transliterated data.}}
\label{table:letter_mappings}
\centering
\resizebox{0.95\textwidth}{!}{%
\begin{tabular}{lp{2.75cm}p{2.75cm}p{2.75cm}p{2.75cm}p{2.75cm}}
 \textbf{L} & \textbf{Algerian} & \textbf{Egyptian} & \textbf{Lebanese} & \textbf{Moroccan} & \textbf{Tunisian} \\
\midrule
$\epsilon$ & \textcolor{black}{e~(270)} \textcolor{black}{a~(211)} \textcolor{black}{o~(36)} \textcolor{black}{u~(18)} \textcolor{black}{i~(17)} \textcolor{black}{d~(12)} \textcolor{black}{'~(10)} \textcolor{black}{sp.~(7)} \textcolor{black}{y~(1)} & \textcolor{black}{a~(387)} \textcolor{black}{e~(211)} \textcolor{black}{o~(112)} \textcolor{black}{i~(17)} \textcolor{black}{sp.~(12)} \textcolor{black}{u~(9)} & \textcolor{black}{a~(340)} \textcolor{black}{e~(135)} \textcolor{black}{i~(18)} \textcolor{black}{o~(14)} \textcolor{black}{u~(12)} \textcolor{black}{sp.~(6)} & \textcolor{black}{a~(209)} \textcolor{black}{e~(188)} \textcolor{black}{i~(72)} \textcolor{black}{o~(58)} \textcolor{black}{u~(33)} \textcolor{black}{d~(14)} \textcolor{black}{sp.~(2)} \textcolor{black}{-~(1)} \textcolor{black}{y~(1)} \textcolor{black}{'~(1)} \textcolor{black}{ue~(1)} & \textcolor{black}{a~(332)} \textcolor{black}{e~(197)} \textcolor{black}{o~(101)} \textcolor{black}{sp.~(16)} \textcolor{black}{i~(15)} \textcolor{black}{d~(14)} \textcolor{black}{ou~(3)} \textcolor{black}{'~(2)} \textcolor{black}{é~(1)} \textcolor{black}{u~(1)} \\
\midrule
sp. & \textcolor{black}{sp.~(17)} & \textcolor{black}{sp.~(20)} & \textcolor{black}{sp.~(7)} & \textcolor{black}{sp.~(9)} & \textcolor{black}{sp.~(23)} \\
\midrule
\AR{ء} & \textbf{$\epsilon$~(14)} \textbf{a~(6)} \textbf{e~(2)} \textbf{2~(2)} \textcolor{black}{2~(1)} \textbf{'a~(1)} \textbf{aa~(1)} & \textcolor{black}{2~(43)} \textbf{a2~(4)} \textbf{'~(2)} & \textcolor{black}{2~(13)} \textbf{i’~(1)} & \textcolor{black}{2~(14)} \textbf{ae~(2)} \textbf{a~(1)} & \textbf{$\epsilon$~(14)} \textbf{e~(7)} \textcolor{black}{2~(3)} \textbf{2~(2)} \textbf{a~(2)} \textbf{a2~(1)} \\
\midrule
\AR{آ} & \textbf{a~(12)} \textbf{e~(2)} & \textbf{a~(17)} \textbf{2a~(7)} \textbf{aa~(1)} & \textbf{e~(9)} \textbf{2e~(3)} \textbf{2a~(1)} \textbf{a~(1)} & \textbf{a~(17)} \textbf{2a~(1)} & \textbf{e~(14)} \textbf{a~(14)} \\
\midrule
\AR{أ} & \textbf{a~(20)} & \textbf{a~(66)} \textcolor{black}{2~(6)} \textbf{2a~(1)} & \textbf{a~(24)} \textcolor{black}{2~(4)} & \textbf{a~(39)} \textbf{2a~(1)} & \textbf{a~(38)} \textbf{e~(20)} \textbf{$\epsilon$~(1)} \\
\midrule
\AR{ؤ} & \textbf{ou2~(7)} \textbf{o~(5)} \textbf{ou~(4)} \textbf{ou'~(2)} \textbf{o2~(1)} & \textbf{o2~(25)} & \textbf{ou2~(9)} \textbf{o2~(3)} \textbf{u2~(2)} & \textbf{ou2~(11)} \textbf{o2~(5)} \textbf{ou~(2)} \textbf{u2~(1)} \textbf{ou-~(1)} & \textbf{ou2~(22)} \textbf{ou~(6)} \textbf{o2~(1)} \\
\midrule
\AR{إ} & \textbf{i~(19)} & \textbf{e~(19)} \textcolor{black}{2~(4)} \textbf{i~(2)} & \textbf{e~(12)} \textbf{i~(10)} \textcolor{black}{2~(2)} \textbf{a~(1)} \textbf{2i~(1)} \textbf{2e~(1)} & \textbf{i~(21)} \textcolor{black}{2~(1)} & \textbf{e~(20)} \textbf{$\epsilon$~(7)} \textbf{i~(3)} \\
\midrule
\AR{ا} & \textcolor{black}{a~(476)} \textcolor{black}{e~(21)} \textbf{$\epsilon$~(9)} \textbf{aa~(5)} \textbf{2~(3)} \textbf{a'~(1)} \textbf{ao~(1)} \textbf{eeee~(1)} \textcolor{black}{è~(1)} \textbf{i~(1)} & \textcolor{black}{a~(686)} \textcolor{black}{e~(60)} \textbf{$\epsilon$~(31)} \textbf{aa~(25)} \textbf{2~(5)} \textbf{i~(4)} & \textcolor{black}{a~(183)} \textcolor{black}{e~(104)} \textbf{$\epsilon$~(24)} \textbf{i~(5)} \textbf{2~(4)} \textbf{ei~(1)} & \textcolor{black}{a~(530)} \textbf{$\epsilon$~(43)} \textbf{i~(17)} \textcolor{black}{e~(3)} & \textcolor{black}{a~(498)} \textcolor{black}{e~(242)} \textbf{$\epsilon$~(40)} \textbf{i~(20)} \textbf{é~(4)} \textbf{ee~(2)} \textbf{y~(1)} \textbf{ia~(1)} \\
\midrule
\AR{ب} & \textcolor{black}{b~(153)} \textcolor{black}{p~(17)} \textbf{bb~(10)} & \textcolor{black}{b~(263)} \textcolor{black}{p~(10)} & \textcolor{black}{b~(71)} \textcolor{black}{p~(14)} & \textcolor{black}{b~(148)} \textcolor{black}{p~(20)} \textbf{ba~(2)} & \textcolor{black}{b~(263)} \textcolor{black}{p~(23)} \\
\midrule
\AR{ة} & \textcolor{black}{a~(249)} \textcolor{black}{at~(9)} \textcolor{black}{et~(6)} \textcolor{black}{e~(1)} & \textcolor{black}{a~(117)} \textcolor{black}{et~(10)} \textbf{ah~(7)} \textbf{t~(5)} \textcolor{black}{at~(4)} \textbf{aa~(1)} & \textcolor{black}{a~(49)} \textcolor{black}{e~(16)} \textbf{t~(7)} \textcolor{black}{eh~(3)} \textbf{i~(1)} & \textcolor{black}{a~(95)} \textcolor{black}{at~(10)} \textbf{t~(2)} & \textcolor{black}{a~(107)} \textcolor{black}{et~(19)} \textbf{i~(3)} \textbf{t~(1)} \textbf{it~(1)} \textcolor{black}{at~(1)} \textbf{ét~(1)} \\
\midrule
\AR{ت} & \textcolor{black}{t~(57)} \textbf{$\epsilon$~(14)} & \textcolor{black}{t~(124)} & \textcolor{black}{t~(57)} \textbf{z~(14)} & \textcolor{black}{t~(74)} & \textcolor{black}{t~(90)} \\
\midrule
\AR{ث} & \textcolor{black}{th~(15)} \textcolor{black}{t~(4)} & \textcolor{black}{th~(23)} \textcolor{black}{s~(17)} \textcolor{black}{t~(9)} & \textcolor{black}{s~(20)} \textcolor{black}{th~(7)} & \textcolor{black}{t~(16)} \textcolor{black}{th~(2)} & \textcolor{black}{th~(24)} \\
\midrule
\AR{ج} & \textcolor{black}{j~(52)} \textcolor{black}{dj~(25)} \textcolor{black}{g~(9)} \textbf{dj'~(1)} & \textcolor{black}{g~(187)} \textcolor{black}{j~(6)} \textbf{sh~(1)} & \textcolor{black}{j~(72)} \textcolor{black}{g~(11)} & \textcolor{black}{j~(38)} & \textcolor{black}{j~(84)} \\
\midrule
\AR{ح} & \textcolor{black}{7~(88)} \textcolor{black}{h~(68)} & \textcolor{black}{7~(100)} \textcolor{black}{h~(24)} & \textcolor{black}{7~(43)} \textcolor{black}{h~(34)} & \textcolor{black}{7~(23)} \textcolor{black}{h~(21)} \textbf{h7~(1)} & \textcolor{black}{7~(129)} \textcolor{black}{h~(57)} \\
\midrule
\AR{خ} & \textcolor{black}{kh~(20)} \textcolor{black}{5~(1)} & \textcolor{black}{5~(17)} \textcolor{black}{kh~(8)} & \textcolor{black}{kh~(16)} \textcolor{black}{5~(12)} & \textcolor{black}{kh~(80)} & \textcolor{black}{5~(29)} \textcolor{black}{kh~(28)} \\
\midrule
\AR{د} & \textcolor{black}{d~(39)} & \textcolor{black}{d~(94)} & \textcolor{black}{d~(49)} \textbf{dd~(11)} \textbf{$\epsilon$~(2)} & \textcolor{black}{d~(56)} & \textcolor{black}{d~(165)} \\
\midrule
\AR{ذ} & \textcolor{black}{d~(16)} \textcolor{black}{dh~(4)} & \textcolor{black}{z~(17)} \textcolor{black}{d~(6)} \textcolor{black}{th~(2)} & \textcolor{black}{z~(14)} & \textcolor{black}{d~(22)} & \textcolor{black}{dh~(25)} \textcolor{black}{th~(4)} \textbf{4~(1)} \\
\midrule
\AR{ر} & \textcolor{black}{r~(220)} \textbf{rr~(15)} & \textcolor{black}{r~(219)} & \textcolor{black}{r~(154)} & \textcolor{black}{r~(322)} \textbf{rr~(16)} & \textcolor{black}{r~(229)} \\
\midrule
\AR{ز} & \textcolor{black}{z~(31)} \textbf{zz~(8)} & \textcolor{black}{z~(24)} & \textcolor{black}{z~(28)} \textbf{s~(13)} \textbf{zz~(1)} & \textcolor{black}{z~(52)} \textbf{s~(1)} \textbf{zz~(1)} &  \\
\midrule
\AR{س} & \textcolor{black}{s~(127)} \textbf{ss~(23)} & \textcolor{black}{s~(174)} & \textcolor{black}{s~(107)} \textbf{ss~(5)} & \textcolor{black}{s~(133)} \textbf{ss~(31)} & \textcolor{black}{s~(166)} \textbf{ss~(14)} \\
\midrule
\AR{ش} & \textcolor{black}{ch~(55)} \textcolor{black}{sh~(2)} \textbf{s~(1)} & \textcolor{black}{sh~(23)} \textcolor{black}{4~(2)} & \textcolor{black}{sh~(20)} \textcolor{black}{ch~(8)} & \textcolor{black}{ch~(60)} \textcolor{black}{sh~(1)} & \textcolor{black}{ch~(51)} \textcolor{black}{sh~(9)} \\
\midrule
\AR{ص} & \textcolor{black}{s~(64)} \textbf{ss~(16)} & \textcolor{black}{s~(49)} & \textcolor{black}{s~(29)} & \textcolor{black}{s~(30)} \textbf{ss~(16)} & \textcolor{black}{s~(55)} \\
\midrule
\AR{ض} & \textcolor{black}{d~(14)} \textcolor{black}{dh~(4)} \textbf{dd~(1)} & \textcolor{black}{d~(25)} & \textcolor{black}{d~(10)} \textbf{t~(4)} & \textcolor{black}{d~(16)} \textcolor{black}{dh~(1)} & \textcolor{black}{dh~(25)} \textcolor{black}{d~(1)} \textbf{4~(1)} \\
\midrule
\AR{ط} & \textcolor{black}{t~(52)} \textcolor{black}{6~(6)} \textbf{tt~(2)} & \textcolor{black}{t~(24)} \textcolor{black}{6~(1)} & \textcolor{black}{t~(27)} \textbf{d~(1)} & \textcolor{black}{t~(65)} & \textcolor{black}{t~(56)} \\
\midrule
\AR{ظ} & \textbf{d~(11)} \textcolor{black}{dh~(6)} \textcolor{black}{th~(1)} & \textcolor{black}{z~(24)} \textcolor{black}{th~(1)} & \textcolor{black}{z~(14)} & \textbf{d~(20)} \textcolor{black}{dh~(2)} \textbf{dd~(1)} & \textcolor{black}{dh~(23)} \textbf{d~(4)} \textbf{4~(1)} \\
\midrule
\AR{ع} & \textcolor{black}{3~(38)} \textbf{a~(2)} & \textcolor{black}{3~(98)} \textbf{'~(1)} & \textcolor{black}{3~(44)} \textbf{a~(11)} \textbf{e~(1)} & \textcolor{black}{3~(97)} \textbf{aea~(1)} \textbf{ee~(1)} \textbf{a~(1)} \textbf{e~(1)} \textbf{ea~(1)} & \textcolor{black}{3~(133)} \textbf{a~(11)} \textbf{aa~(7)} \textbf{ee~(1)} \\
\midrule
\AR{غ} & \textcolor{black}{gh~(19)} & \textcolor{black}{gh~(14)} \textcolor{black}{8~(5)} \textcolor{black}{3'~(5)} & \textcolor{black}{gh~(10)} \textcolor{black}{8~(5)} & \textcolor{black}{gh~(18)} & \textcolor{black}{gh~(21)} \textcolor{black}{8~(9)} \\
\midrule
\AR{ف} & \textcolor{black}{f~(53)} \textcolor{black}{v~(28)} & \textcolor{black}{f~(53)} \textcolor{black}{v~(21)} \textbf{ph~(1)} & \textcolor{black}{f~(56)} & \textcolor{black}{f~(60)} \textcolor{black}{v~(20)} & \textcolor{black}{f~(62)} \textcolor{black}{v~(29)} \\
\midrule
\AR{ق} & \textcolor{black}{9~(65)} \textcolor{black}{q~(22)} \textbf{k~(14)} \textcolor{black}{g~(8)} \textbf{9~(7)} \textbf{gh~(2)} & \textcolor{black}{2~(23)} \textcolor{black}{q~(21)} \textbf{k~(5)} \textbf{'~(1)} & \textcolor{black}{2~(30)} \textbf{k~(9)} \textbf{a~(9)} \textcolor{black}{q~(1)} & \textcolor{black}{9~(53)} \textbf{9~(30)} \textcolor{black}{q~(29)} \textcolor{black}{g~(8)} \textbf{que~(2)} \textbf{qu~(2)} & \textcolor{black}{9~(19)} \textcolor{black}{g~(8)} \textbf{k~(2)} \\
\midrule
\AR{ك} & \textcolor{black}{k~(172)} \textbf{c~(18)} & \textcolor{black}{k~(71)} \textbf{q~(4)} & \textcolor{black}{k~(21)} \textbf{kk~(7)} & \textcolor{black}{k~(49)} \textcolor{black}{g~(43)} \textbf{c~(32)} \textbf{q~(2)} \textbf{gu~(2)} & \textcolor{black}{k~(145)} \textbf{c~(17)} \textbf{kk~(7)} \textbf{9~(1)} \textbf{cc~(1)} \\
\midrule
\AR{ل} & \textcolor{black}{l~(196)} \textbf{ll~(9)} & \textcolor{black}{l~(317)} \textbf{ll~(1)} & \textcolor{black}{l~(147)} & \textcolor{black}{l~(274)} \textbf{ll~(31)} & \textcolor{black}{l~(257)} \textbf{ll~(28)} \\
\midrule
\AR{م} & \textcolor{black}{m~(198)} \textbf{mm~(10)} & \textcolor{black}{m~(332)} \textbf{mm~(17)} & \textcolor{black}{m~(195)} & \textcolor{black}{m~(211)} \textbf{mm~(13)} & \textcolor{black}{m~(280)} \textbf{mm~(19)} \\
\midrule
\AR{ن} & \textcolor{black}{n~(130)} \textbf{nn~(13)} & \textcolor{black}{n~(387)} & \textcolor{black}{n~(57)} & \textcolor{black}{n~(129)} & \textcolor{black}{n~(186)} \textbf{nn~(10)} \\
\midrule
\AR{ه} & \textcolor{black}{h~(35)} \textbf{8~(5)} \textcolor{black}{ah~(1)} & \textcolor{black}{h~(46)} \textbf{t~(3)} \textbf{$\epsilon$~(1)} & \textcolor{black}{h~(28)} & \textcolor{black}{h~(45)} \textbf{8~(2)} & \textcolor{black}{h~(110)} \textbf{hi~(3)} \textbf{$\epsilon$~(1)} \textcolor{black}{eh~(1)} \\
\midrule
\AR{و} & \textcolor{black}{ou~(104)} \textcolor{black}{w~(52)} \textcolor{black}{o~(27)} \textbf{ww~(4)} \textcolor{black}{u~(3)} & \textcolor{black}{w~(134)} \textcolor{black}{o~(69)} \textcolor{black}{oo~(45)} \textcolor{black}{u~(12)} \textcolor{black}{ou~(4)} \textbf{ww~(2)} \textbf{uu~(1)} & \textcolor{black}{w~(117)} \textcolor{black}{o~(32)} \textcolor{black}{ou~(22)} \textcolor{black}{u~(1)} & \textcolor{black}{w~(63)} \textcolor{black}{ou~(46)} \textcolor{black}{o~(29)} \textcolor{black}{u~(4)} \textbf{$\epsilon$~(1)} & \textcolor{black}{ou~(166)} \textcolor{black}{w~(112)} \textcolor{black}{o~(8)} \textbf{ow~(1)} \\
\midrule
\AR{ى} & \textcolor{black}{a~(37)} & \textcolor{black}{a~(24)} \textbf{ah~(1)} & \textcolor{black}{a~(13)} \textbf{e~(1)} & \textcolor{black}{a~(21)} & \textcolor{black}{a~(23)} \\
\midrule
\AR{ي} & \textcolor{black}{i~(172)} \textcolor{black}{y~(90)} \textcolor{black}{a~(1)} \textbf{ya~(1)} & \textcolor{black}{y~(119)} \textcolor{black}{i~(54)} \textbf{e~(48)} \textcolor{black}{ee~(45)} \textbf{ye~(20)} \textbf{yy~(2)} \textcolor{black}{ei~(1)} \textbf{ie~(1)} \textbf{yee~(1)} \textbf{ey~(1)} \textbf{ii~(1)} & \textcolor{black}{i~(134)} \textbf{e~(32)} \textcolor{black}{y~(21)} \textbf{yy~(7)} \textcolor{black}{ei~(6)} \textcolor{black}{ee~(3)} & \textcolor{black}{i~(208)} \textcolor{black}{y~(89)} \textbf{yy~(9)} \textbf{iy~(8)} \textcolor{black}{ei~(2)} \textbf{ya~(2)} \textbf{$\epsilon$~(1)} \textbf{ui~(1)} & \textcolor{black}{i~(340)} \textcolor{black}{y~(186)} \textbf{ya~(25)} \textbf{yy~(6)} \textbf{$\epsilon$~(3)} \textbf{e~(2)} \textbf{ay~(1)} \textcolor{black}{é~(1)} \textbf{yi~(1)} \\
\midrule
\FR{چ} & \textcolor{black}{g~(8)} \textcolor{black}{j~(1)} \textbf{dj~(1)} & \textcolor{black}{g~(24)} \textbf{sh~(1)} &  &  &  \\
\midrule
\textkurdish{ڜ} & \textcolor{black}{ch~(64)} \textcolor{black}{tch~(8)} &  &  &  &  \\
\midrule
\textkurdish{ڤ} & \textbf{g~(32)} \textbf{gh~(3)} \textbf{9~(2)} \textbf{f~(1)} & \textcolor{black}{v~(24)} & \textcolor{black}{v~(14)} & \textcolor{black}{v~(21)} \textbf{9~(1)} & \textbf{g~(19)} \\
\midrule
\AR{ئ} &  & \textcolor{black}{2~(24)} \textbf{e2~(16)} \textbf{ee~(7)} \textbf{a'~(2)} \textbf{i2~(1)} & \textcolor{black}{2~(13)} \textbf{e~(1)} & \textbf{2i~(17)} \textcolor{black}{2~(17)} \textbf{a~(4)} \textbf{i~(2)} \textbf{e~(1)} \textbf{$\epsilon$~(1)} & \textcolor{black}{2~(25)} \textbf{a~(2)} \textbf{ay~(1)} \textbf{ai~(1)} \textbf{ae~(1)} \\
\midrule
\textkurdish{ڭ} &  &  &  & \textbf{gu~(19)} \textbf{g~(2)} &  \\
\midrule
\textkurdish{ڥ} &  &  &  &  & \textcolor{black}{v~(30)} \\
\bottomrule
\end{tabular}%
}
\end{table*}